\documentclass[journal,twoside, a4paper]{IEEEtran}
\usepackage{multicol}
\usepackage{makecell}
\usepackage{color}
\usepackage{amssymb}
\usepackage{rotating}
\usepackage[T1]{fontenc}
\usepackage{lscape}

\usepackage{xcolor}

\usepackage[ruled,vlined]{algorithm2e}

\usepackage{multirow}
\usepackage{enumitem}

\usepackage{hyperref}

\usepackage{float}
\usepackage{algorithmic}

\usepackage{xcolor}

\usepackage{amsmath}  
\usepackage{bm}

\def\BibTeX{{\rm B\kern-.05em{\sc i\kern-.025em b}\kern-.08em
    T\kern-.1667em\lower.7ex\hbox{E}\kern-.125emX}}

\ifCLASSINFOpdf
\else
 
\fi

\usepackage[a4paper, total={180mm,257mm}]{geometry}

\begin{document}

\title{Leveraging Transfer Learning and Mobile-enabled Convolutional Neural Networks for Improved Arabic Handwritten Character Recognition}

\author{\IEEEauthorblockN{
Mohsine El Khayati\IEEEauthorrefmark{1}, 
Ayyad Maafiri\IEEEauthorrefmark{2}, 
Yassine Himeur\IEEEauthorrefmark{3},  
Hamzah Ali Alkhazaleh\IEEEauthorrefmark{3},  
Shadi Atalla\IEEEauthorrefmark{3},  
and Wathiq Mansoor\IEEEauthorrefmark{3}
}\\
\IEEEauthorblockA{\IEEEauthorrefmark{1} 
Systems Theory and Informatics Laboratory, University Moulay Ismail, Meknes, Morocco; 
Email: m.elkhayati@umi.ac.ma}\\
\IEEEauthorblockA{\IEEEauthorrefmark{2} 
LMC, Polydisciplinary Faculty of Safi, Cadi Ayyad University, Morocco; 
Email: a.maafiri@uca.ac.ma}\\
\IEEEauthorblockA{\IEEEauthorrefmark{3} 
College of Engineering and Information Technology, University of Dubai, Dubai, UAE; 
Emails: yhimeur@ud.ac.ae; halkhazaleh@ud.ac.ae; satalla@ud.ac.ae; wmansoor@ud.ac.ae}
}

\maketitle
\thispagestyle{empty}
\pagestyle{empty}

\begin{abstract}
The study explores the integration of transfer learning (TL) with mobile-enabled convolutional neural networks (MbNets) to enhance Arabic Handwritten Character Recognition (AHCR). Addressing challenges like extensive computational requirements and dataset scarcity, this research evaluates three TL strategies—full fine-tuning, partial fine-tuning, and training from scratch—using four lightweight MbNets: MobileNet, SqueezeNet, MnasNet, and ShuffleNet. Experiments were conducted on three benchmark datasets: AHCD, HIJJA, and IFHCDB. MobileNet emerged as the top-performing model, consistently achieving superior accuracy, robustness, and efficiency, with ShuffleNet excelling in generalization, particularly under full fine-tuning. The IFHCDB dataset yielded the highest results, with 99\% accuracy using MnasNet under full fine-tuning, highlighting its suitability for robust character recognition. The AHCD dataset achieved competitive accuracy (97\%) with ShuffleNet, while HIJJA posed significant challenges due to its variability, achieving a peak accuracy of 92\% with ShuffleNet. Notably, full fine-tuning demonstrated the best overall performance, balancing accuracy and convergence speed, while partial fine-tuning underperformed across metrics. These findings underscore the potential of combining TL and MbNets for resource-efficient AHCR, paving the way for further optimizations and broader applications. Future work will explore architectural modifications, in-depth dataset features analysis, data augmentation, and advanced sensitivity analysis to enhance model robustness and generalizability.
\end{abstract}


\section{Introduction}  \label{intro}
\label{Introduction}

Arabic Handwriting Recognition (AHR) is a complex and evolving research domain focused on transforming handwritten text from image format into a machine-readable form \cite{el2024cnn}. The challenge stems from the intricate attributes of handwritten Arabic script \cite{1}, including its cursive nature, the presence of diacritics, artistic nuances, and structural complexities such as over-traces, overlaps, touching elements, and vertical ligatures \cite{alshehri2024deepahr}. Recent advancements in AHR have been driven largely by the adoption of sophisticated deep learning techniques, particularly Convolutional Neural Networks (ConvNets) \cite{mahdi2024advancing,rouabhi2024conv}. ConvNets have demonstrated exceptional performance, especially in Arabic Handwritten Character Recognition (AHCR) \cite{kheddar2024automatic}.

In recent years, the development of deeper ConvNet models has surged, delivering outstanding results in computer vision applications \cite{copiaco2023innovative,himeur2023face, Anand2025}. However, these deep models come with significant computational demands due to their large parameter counts, which can reach up to 3.9 billion \cite{2}, and their high Multiply Accumulation Operations (MACs) and Floating Point Operations (FLOPs), which can go as high as 2800 GFLOPs \cite{3}. Moreover, deep ConvNets are prone to overfitting and require extensive datasets to perform optimally, necessitating ongoing research to address these challenges.

Transfer Learning (TL) and Mobile-enabled ConvNet architectures (MbNets) have emerged as pivotal solutions to these issues \cite{himeur2023video,kheddar2023deep}. TL effectively mitigates data scarcity \cite{4,5} by leveraging pre-trained models from related tasks, reducing training time, and eliminating the need to develop models from scratch. Meanwhile, MbNets provide efficient and lightweight architectures that significantly impact memory usage, training/inference times, and deployability \cite{himeur2022deep}.

This study explores the efficiency and effectiveness of TL and MbNets in the context of AHCR, moving beyond the traditional use of heavyweight models such as VGGNet \cite{Shiferaw2025}, Inception V3 \cite{6,7}, ResNet \cite{6,7,8, Shiferaw2025}, and AlexNet \cite{9,10}, which are characterized by substantial parameter loads ranging from 27 million to 143 million. Specifically, we investigate three distinct TL strategies: (i) Training From Scratch (TFS), (ii) Training as a Fixed Feature Extractor (TFE), and (iii) Training as a Weight Initializer (TWI). We assess these strategies in combination with four MbNets—MobileNet, SqueezeNet, MnasNet, and ShuffleNet—on three benchmark Arabic datasets: AHCD \cite{11}, Hijja \cite{12}, and IFHCDB \cite{13}. This comprehensive evaluation provides valuable insights and actionable recommendations. 

Although the proposed models are trained and evaluated on isolated Arabic characters, such systems have immediate and tangible applications \cite{ElKhayati2024}. Examples include mobile OCR tools for digitizing handwritten forms where characters are written separately, automated grading of quiz answers on sheets in educational or administrative contexts, and teaching aids for character writing in early school programs to support literacy development. They can also be used in rapid data-entry systems for field surveys or postal code recognition, where isolated character input is common. The adoption of mobile-friendly architectures ensures these applications can run efficiently on resource-constrained devices, enabling real-time operation without reliance on high-performance hardware.

The main contributions of this work are as follows:

\begin{itemize}
\item Pioneering the fusion of transfer learning (TL) with mobile-enabled architectures in Arabic Handwritten Recognition (AHR), introducing an innovative approach in the literature. This combination addresses the need for efficient yet accurate solutions in AHR, marking a significant advancement in the field.
\item Uncovering the potential of lesser-explored models like MnasNet, ShuffleNet, and SqueezeNet in AHR. Through empirical analysis, these models are showcased for their efficacy and advantages, expanding the available toolkit for researchers in the field.
\item	Conducting a meticulous array of experiments to systematically evaluate diverse models, TL strategies, and datasets in AHR. This rigorous evaluation process extracts pertinent conclusions and insights, providing valuable guidance for researchers in the area of AHR. Through this effort, we contribute to the ongoing evolution and enhancement of AHR methodologies, fostering innovation in the field.
\item	Demonstrating superior performance with reduced complexity in AHCR, challenging the prevailing trend towards deeper, more complex models. By achieving higher accuracies while maintaining lower computational demands, this work opens avenues for refining practices in AHCR, emphasizing efficiency and effectiveness in model design and deployment.
\end{itemize}

The remainder of this paper is organized as follows: Section 2 defines key terms, including the features of Arabic handwritten characters, Convolutional Neural Networks (ConvNets), and Transfer Learning (TL). Section 3 provides a review of relevant literature, with a particular focus on the applications of TL in Arabic Handwriting Recognition (AHR). Section 4 outlines the experimental strategies, models, and datasets employed in this study. Section 5 presents and analyzes the experimental results, offering conclusions and actionable recommendations. Finally, Section 6 concludes the paper by summarizing the findings and discussing directions for future research.

\section{Background}

\subsection{Arabic Handwritten Characters}
The Arabic script is inherently cursive, written from right to left, and comprises 28 characters, of which 22 are fully cursive and 6 are semi-cursive. The shape of each character varies depending on its position within a word—whether isolated, initial, medial, or final. Additionally, Arabic script incorporates diacritical marks, including points and signs. Diacritical points are essential for distinguishing characters, as 16 of them feature points above or below the main character's body. In certain cases, characters with similar base shapes are differentiated solely by the placement or number of diacritical points. Meanwhile, diacritical signs, akin to vowels in English, influence pronunciation and word meanings. However, most Arabic texts omit these signs, as native readers can infer pronunciation from context.

\begin{figure*}[t]
  \centering
  \includegraphics[width=0.75\textwidth]{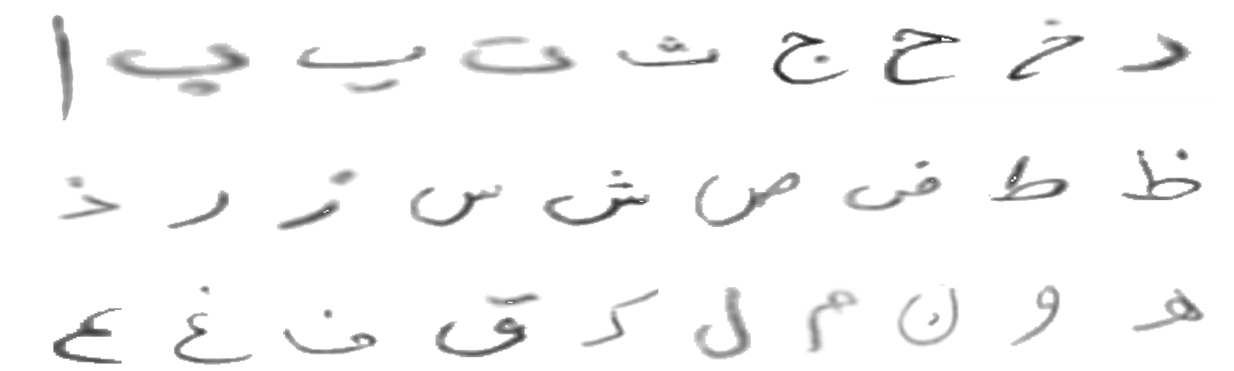}
  \caption{The forms of isolated Arabic handwritten characters, sourced from the IFHCDB dataset.}
  \label{fig1}
\end{figure*}

The unique characteristics of Arabic script, including its dynamic shapes and reliance on diacritical elements, add considerable complexity to Arabic Handwritten Character Recognition (AHCR). These factors make AHCR one of the most challenging problems in handwriting recognition algorithms. Figure \ref{fig1} illustrates the isolated forms of the 28 Arabic characters, sourced from the IFHCDB dataset.

\subsection{Convolutional Neural Networks}
Deep learning has emerged as one of the most transformative advancements in computer science, particularly due to its hierarchical structures that enable the extraction of high-level features from data \cite{14}. A key milestone in this field was the development of Convolutional Neural Networks (ConvNets), a type of deep feedforward artificial neural network inspired by biological processes observed in the animal cortex.

ConvNets have gained widespread recognition for their ability to solve complex problems, often outperforming other learning methods. Their strength lies in their capacity to handle high-dimensional inputs, capture intricate non-linear relationships, and autonomously identify and extract essential invariant features from data \cite{15,16}. Furthermore, ConvNets leverage shared weights in convolutional layers, reducing the number of parameters and improving overall performance \cite{17}.


\subsection{Transfer Learning}
Transfer Learning (TL) is a machine learning technique that leverages pre-trained models, originally trained on large datasets, for smaller, related tasks. The parameters learned during the initial training phase are reused for retraining or testing on the new task \cite{kheddar2025breathai,kerdjidj2025exploring}. Developing deep models from scratch is computationally intensive and requires vast datasets to achieve convergence. TL addresses these challenges by enabling knowledge transfer from well-trained models, such as those trained on the ImageNet dataset, to domains with limited data \cite{bechar2025federated,sohail2025advancing}. This approach accelerates convergence, enhances generalization \cite{18}, and reduces both training time and computational costs.

\begin{figure*}[t!]
\begin{center}
\includegraphics[width=0.75\textwidth]{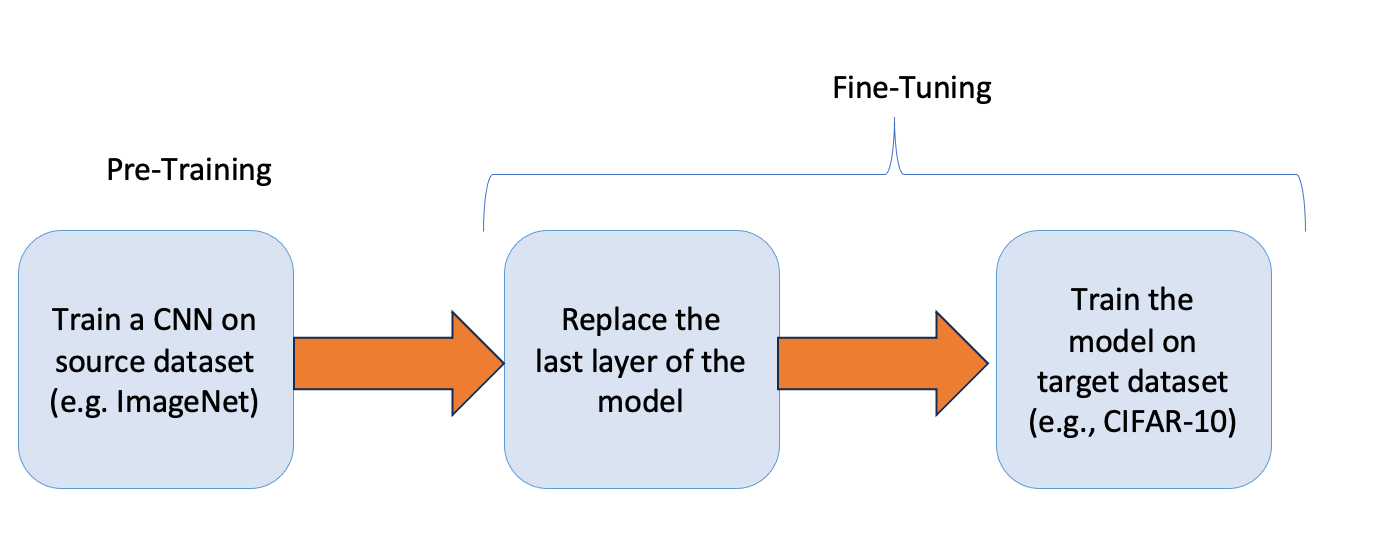} 
\caption{Illustrative diagram depicting the concept of transfer learning.}
\label{fig2}
\end{center}
\end{figure*}

Figure \ref{fig2} provides a visual representation of the TL concept. Various TL methods have been proposed, which can be categorized into four main types \cite{4}:
\begin{itemize}
    \item \textbf{Mapping-based approaches}: Transform instances from both the source and target domains into a new shared feature space \cite{19,20}.
    \item \textbf{Adversarial-based approaches}: Use adversarial techniques to identify transferable representations between the source and target domains \cite{21,22}.
    \item \textbf{Instance-based approaches}: Select partial instances from the source domain to supplement the target training set \cite{23,24}.
    \item \textbf{Network-based approaches}: Reuse parts of pre-trained network models from the source domain, which is the approach adopted in this study \cite{25,26}.
\end{itemize}

\section{Related Works}

Transfer Learning (TL) has recently emerged as a promising approach in Arabic Handwriting Recognition (AHR). Researchers have leveraged pre-trained Convolutional Neural Network (ConvNet) models, initially trained on extensive datasets such as ImageNet, and fine-tuned them on Arabic handwriting datasets. This section provides an overview of TL techniques employed in AHR.

Boufenar et al. \cite{9} explored three TL strategies: (i) training from scratch, (ii) using a ConvNet as a feature extractor, and (iii) fine-tuning the ConvNet. Their experiments, conducted using an 8-layer ConvNet based on AlexNet, revealed an intriguing finding—training from scratch outperformed TL approaches, achieving near-perfect accuracy under specific conditions. This raises questions about the effectiveness of TL in Arabic Handwritten Character Recognition (AHCR), particularly given the stark performance gap favoring training from scratch.

Soumia et al. \cite{6} compared traditional machine learning with a Support Vector Machine (SVM) classifier, TL with ResNet, Inception V3, and VGG16, and a custom ConvNet trained from scratch. Their custom ConvNet achieved the highest accuracy, with 94.7\% on the OHACDB-28 dataset, 98.3\% on OHACDB-40, and 95.2\% on AIA9K. These findings align with \cite{9}, further suggesting that training from scratch may surpass TL in specific contexts.

Awni et al. \cite{8} examined two TL strategies using the pre-trained ResNet18 model from ImageNet. The first approach assessed TL's effectiveness for cross-domain tasks on the IFN/ENIT dataset, demonstrating modest performance improvements—0.37\% with full fine-tuning and 0.7\% by freezing earlier convolutional layers. The second strategy trained ResNet18 on a separate AHR dataset (Alex-U) and transferred the knowledge to the IFN/ENIT dataset. Remarkably, this approach yielded a 2.64\% improvement with fine-tuning and a significant 3.12\% gain when freezing earlier layers. These results underscore the potential of TL for related tasks but differ from findings in \cite{9}, highlighting the influence of application-specific factors.

Building on the second strategy in \cite{8}, Noubigh et al. \cite{27} applied TL to improve ConvNet-BLSTM-CTC models on the KHATT and AHTID datasets. By transferring knowledge from a mixed-font printed text dataset (P-KHATT), they significantly reduced Word Error Rate (WER) and Character Error Rate (CER), reinforcing TL's advantage over training from scratch.

Masruroh et al. \cite{7} employed TL with several ConvNet architectures, including ResNet, DenseNet, VGG16, VGG19, InceptionV3, and MobileNet. To mitigate overfitting, they applied data augmentation and a 50\% dropout rate. Among the models, VGG16, trained with the Adam optimizer and a learning rate of 0.0001, delivered the best performance on the Hijja and AHCD datasets, favoring heavyweight models for TL in this context.

Arif and Poruran \cite{10} developed two TL-based architectures, OCR-AlexNet and OCR-GoogleNet, by adapting AlexNet and GoogleNet. OCR-AlexNet retained the initial layer weights of AlexNet while modifying the final three layers, while OCR-GoogleNet fine-tuned only four inception modules. These models achieved average accuracies of 96.3\% and 94.7\%, respectively, on the IFHCDB dataset, further supporting the preference for heavyweight models.

Balaha et al. \cite{28} combined TL with genetic algorithms to optimize parameters and hyperparameters, introducing 14 distinct ConvNet architectures. Their highest accuracy on the HMDB dataset was 92.88\%. Meanwhile, Alheraki et al. \cite{29} employed TL with EfficientNetV0 on the Hijja, AHCD, and Hijja-AHCD datasets, achieving accuracies of 86\%, 97\%, and 90\%, respectively. As one of the few studies emphasizing lightweight models, their work aligns closely with the focus of this study.

Most prior research has focused on heavyweight models like VGG16 \cite{Panigrahi2025}, VGG19, Inception V3 \cite{6,7}, ResNet \cite{6,7,8}, and AlexNet \cite{9,10}, known for their substantial parameter counts and Floating Point Operations (FLOPs). While exceptions exist, such as EfficientNet \cite{29} and MobileNet \cite{7}, there has been limited exploration of lightweight models for AHCR.

In this study, we investigate the effectiveness of TL with four Mobile-enabled ConvNet architectures (MbNets) for AHCR. To the best of our knowledge, this is the first exploration of SqueezeNet, MnasNet, and ShuffleNet in this context. Our research not only fills a critical gap in the literature but also broadens the scope of AHCR by emphasizing the potential of lightweight models in conjunction with TL.

\section{Method}
\begin{figure*}[t!]
\begin{center}
\includegraphics[width=0.85\textwidth]{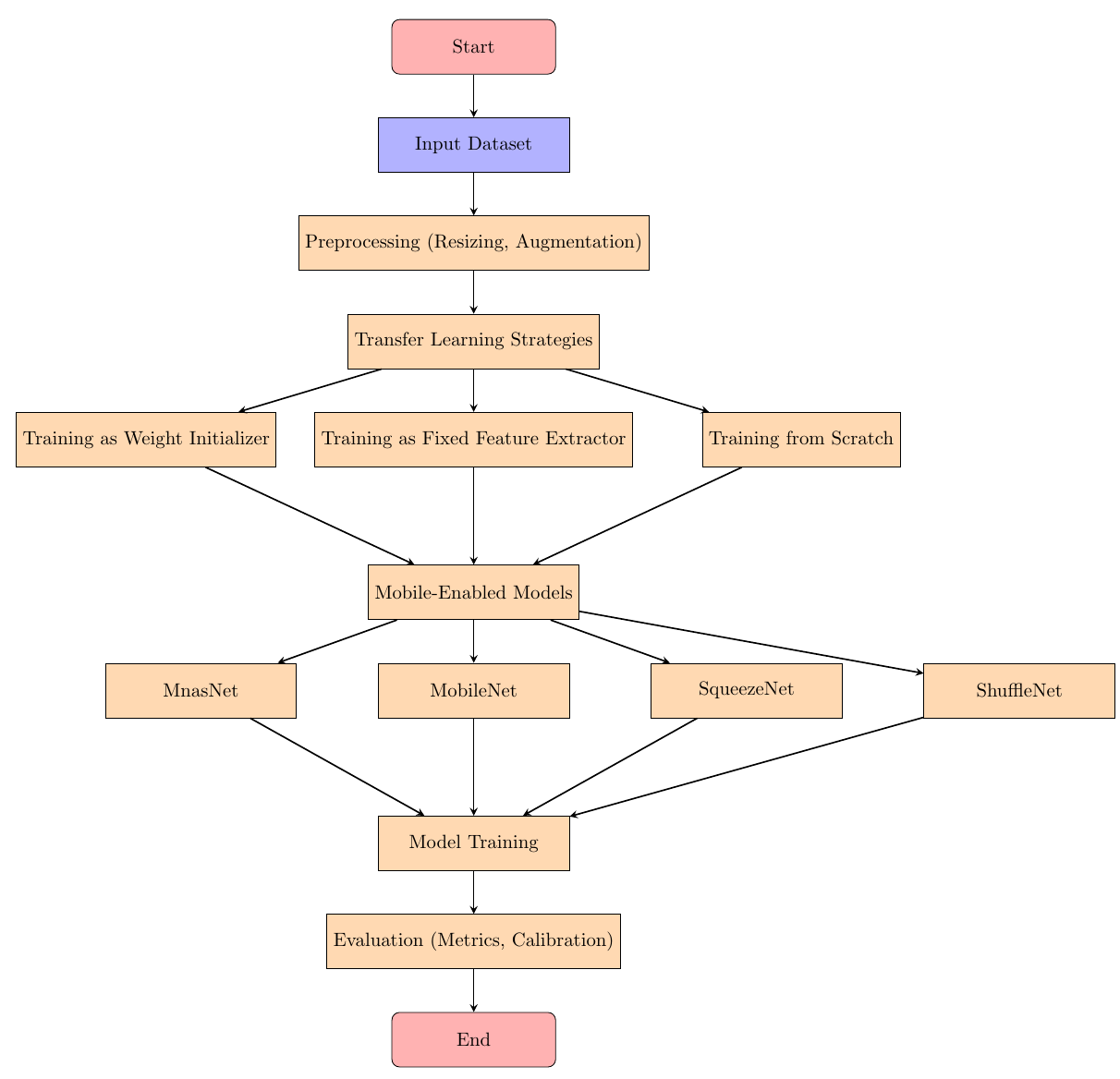} 
\caption{\textcolor{black}{Flowchart of the proposed method for Arabic handwritten character recognition using transfer learning and mobile-enabled CNNs.}}
\label{fig00}
\end{center}
\end{figure*}

This section details the Transfer Learning (TL) strategies employed, the Mobile-enabled Convolutional Neural Network (MbNet) models utilized, the datasets integrated into our research, and the training configurations. Figure \ref{fig00} illustrates the flowchart of the proposed method for Arabic Handwritten Character Recognition (AHCR) using TL and lightweight MbNet models.

\subsection{Transfer Learning Strategies Employed}

In this study, we explore three distinct TL strategies, adapting each model's output layer to match the labels of our training datasets while maintaining consistency across all other layers. Hyperparameters are kept uniform throughout the experiments, as outlined in the Experimental Setup section. The three TL strategies are as follows:

\begin{itemize}
    \item \textbf{Training as Weight Initializer (TWI)}: In this approach, both the model's architecture and pre-trained weights from ImageNet are transferred. Training begins with these learned weights, which are then fine-tuned during the training process to adapt to the new task.
    \item \textbf{Training as a Fixed Feature Extractor (TFE)}: Similar to TWI, this strategy transfers the architecture and pre-trained weights from ImageNet. However, in this case, the feature extraction layers are frozen, and only the classification layers are fine-tuned during training.
    \item \textbf{Training from Scratch (TFS)}: Here, only the model's architecture is utilized without any pre-trained weights. The model is trained entirely from scratch with random weight initialization.
\end{itemize}

\subsection{Studied Models} \label{sec:models}
Top-performing models on ImageNet reach up to 3.9 billion parameters and 2800 GFLOPs \cite{2,3,30,31}, offering high accuracy at the cost of significant computational demand. This limits their use on resource-constrained devices. In contrast, lightweight models, designed for efficiency with fewer parameters and lower FLOPs, have gained traction for mobile and embedded applications \cite{32,33,34,35,36,37,38}. They reduce training time, bandwidth usage, and inference latency while being easier to deploy.

In this study, we selected four lightweight models—MobileNet, ShuffleNet, MnasNet, and SqueezeNet—due to their high efficiency and strong suitability for deployment in resource-constrained environments. These architectures incorporate advanced design strategies such as depthwise separable convolutions, inverted residual blocks, channel shuffling, and Fire modules, all aimed at significantly reducing computational cost and memory usage without compromising accuracy. As illustrated in Table \ref{tab:model_comparison}, these models exhibit substantially lower FLOPs and parameter counts compared to both state-of-the-art ImageNet models and commonly used architectures in Arabic Handwritten Recognition (AHR). While high-resource models such as ViT and CoAtNet are included in the comparison to emphasize the large differences in model size and computational demand, it is important to note that these models are not directly applied to AHR tasks. This comparison serves to validate our model selection strategy, which prioritizes minimal complexity while maintaining competitive performance.

The original dataset split (80\% training / 20\% testing) was predefined by the dataset creators. Within the training set, we applied 5-fold cross-validation, where the training data is split into five subsets. In each fold, four subsets (80\%) were used for training and one (20\%) for validation. The model with the best validation accuracy across folds was selected.

\begin{itemize}
    \item \textbf{MnasNet} \cite{34} uses a latency-aware architecture search, directly optimizing for mobile inference time rather than relying solely on FLOPs.
    \item \textbf{MobileNet} \cite{33,37,38} evolves through three generations, introducing depthwise convolutions, inverted residuals, and NAS-based optimizations for better accuracy and speed trade-offs.
    \item \textbf{SqueezeNet} \cite{32} achieves AlexNet-level accuracy with 50× fewer parameters using 1×1 filters and Fire modules, enabling model sizes as small as 0.5 MB.
    \item \textbf{ShuffleNet} \cite{36} combines group convolutions, channel shuffle, and split-merge operations to achieve low-latency and high-accuracy performance under tight FLOPs constraints.
\end{itemize}

\color{black}
\begin{table*}[t]
\caption{Comparison of our models with those in the literature adopted for Arabic handwriting recognition and some of the top-performing models on ImageNet, considering the number of parameters and GFLOPs.}
\label{tab:model_comparison}
\centering
\begin{tabular}{p{7cm}p{6cm}p{3cm}}
\hline
Models used for AHR in the literature & Parameters in Millions & GFLOPs \\
\hline
\multicolumn{3}{c}{Models used for AHR in the literature} \\
\hline
AlexNet \cite{9,10} & 61 & 0.72 \\
ResNet50 \cite{6,7,8} & 25 & 4 \\
VGG16 \cite{6,7} & 138 & 16 \\
VGG19 \cite{7} & 143 & 20 \\
Inception V3 \cite{6,7} & 27 & 6 \\
\hline
\multicolumn{3}{c}{Top performing models on ImageNet} \\
\hline
CoAtNet-7 \cite{30} & 2440 & 2586 \\
CoAtNet-6 \cite{30} & 1470 & 1521 \\
ViT-G/14 \cite{3} & 1843 & 2859.9 \\
DaViT-G \cite{31} & 1437 & 1038 \\
DaViT-H \cite{31} & 362 & 334 \\
\hline
\multicolumn{3}{c}{The models utilized in this work} \\
\hline
MobileNetV3\_small weights \cite{33} & 2.5 & 0.02 \\
ShuffleNet\_V2\_X0\_5 \cite{35} & 1.4 & 0.013 \\
MnasNet0\_5 \cite{34} & 2.2 & 0.31 \\
SqueezeNet1\_1 \cite{32} & 1.2 & 0.36 \\
\hline
\end{tabular}
\end{table*}

\subsection{Employed Datasets}

The experiments in this study are conducted on three widely recognized benchmark datasets for Arabic handwritten characters: the Arabic Handwritten Character Dataset (AHCD), the Hijja dataset, and the Isolated Farsi Handwritten Character Dataset (IFHCDB). Among these, AHCD is the most commonly used in the literature, while the Hijja dataset has gained traction in recent years but remains less extensively utilized compared to AHCD. IFHCDB, primarily a dataset for Farsi characters, is used less frequently for Arabic character recognition tasks. However, as Farsi script is derived from the Arabic alphabet, it includes all 28 Arabic characters. Table \ref{tab:database_summary} summarizes the key features of these datasets.

\subsubsection{AHCD}

The Arabic Handwritten Character Dataset (AHCD) \cite{11} comprises 16,800 handwritten character samples contributed by 60 participants, aged between 19 and 40. Among the participants, 90\% are right-handed. The dataset is divided into a training set containing 13,440 characters (480 samples per class) and a test set with 3,360 characters (120 samples per class). AHCD is one of the most extensively used datasets for Arabic Handwritten Character Recognition (AHCR) due to its balanced nature and comprehensive representation.

\subsubsection{Hijja}

The Hijja dataset, introduced by Altwaijry and Al-Turaiki \cite{12}, is a publicly available collection of individual Arabic handwritten characters, collected from children aged 7 to 12. It is structured into 108 classes, representing the different shapes of Arabic letters: initial, medial, final, and isolated forms. For this study, we consolidated all shapes of each character into a single class, resulting in a simplified dataset of 28 classes. Each character image in Hijja is $32\times32$ pixels in size, and the dataset includes a total of 47,434 images. The Hijja dataset is particularly noteworthy for its diverse representation of characters across different shapes.

\subsubsection{IFHCDB}

The Isolated Farsi Handwritten Character Dataset (IFHCDB) \cite{13} is primarily a Farsi character dataset but is frequently used for isolated Arabic character recognition due to its inclusion of all 28 Arabic characters. With a total of 52,380 handwritten isolated character samples and an additional 17,740 digit samples, IFHCDB is the most extensive dataset for Arabic characters to date. Approximately 97\% of the dataset consists of Arabic characters. 

A significant challenge with IFHCDB is its imbalanced distribution of samples, where some characters are heavily represented while others have relatively few examples. This imbalance can pose challenges for model training, requiring techniques such as data augmentation or weighted loss functions to ensure fair representation across all classes.

\begin{table*}[t]
\centering
\small
\caption{Summary of characteristics of the datasets used in the experiments.}
\label{tab:database_summary}
\begin{tabular}{p{1.8cm}|p{1.8cm}|p{1.8cm}|p{1.8cm}|p{1.8cm}|p{3cm}|p{3cm}}
\hline
Dataset & Dataset size & Train set & Test set & Images size & Content & Availability \\
\hline
IFHCDB \cite{10} & 70,120 &  36,017 (75\%)& 12,440 (25\%) & $77\times95$ & Character and digits & Non-commercial use \\
AHCD \cite{14} & 16,800 & 13,440 (80\%) & 3,360 (20\%)& $32\times32$ & Characters in isolated form & Yes \\
Hijja \cite{21} & 47,434 & 37,933 (80\%) & 9,500 (20\%) & $32\times32$ & Characters in different forms & Yes \\
\hline
\end{tabular}
\end{table*}

\color{black}
Algorithm \ref{alg:proposed_method} provides a comprehensive overview of the proposed method for Arabic Handwritten Character Recognition (AHCR), leveraging lightweight Mobile-enabled Convolutional Neural Networks (MbNets) and Transfer Learning (TL) strategies. It begins with data preparation, including image resizing, augmentation, and dataset splitting into training, validation, and testing subsets. The algorithm then outlines three TL strategies: Training from Scratch (TFS), Training as a Fixed Feature Extractor (TFE), and Training as a Weight Initializer (TWI), highlighting their respective weight initialization and fine-tuning approaches. The model training process incorporates a cross-entropy loss function, gradient-based optimization, and cross-validation to ensure robust learning. Following training, the models are evaluated using various metrics such as accuracy, precision, recall, F1-score, calibration error (ECE), and sensitivity to noise, providing a thorough assessment of performance and robustness. The final step identifies the best-performing model and strategy for each dataset and saves the optimized weights for deployment, offering an efficient and adaptable framework for AHCR across diverse datasets and conditions.

\begin{algorithm*}[t!]
\SetAlgoLined
\caption{Proposed Method for Arabic Handwritten Character Recognition (AHCR)}
\label{alg:proposed_method}
\KwIn{
    Pre-trained CNN architectures (MobileNet, ShuffleNet, MnasNet, SqueezeNet), Arabic handwriting datasets (AHCD, Hijja, IFHCDB), transfer learning strategies (TFS, TFE, TWI).
}
\KwOut{
    Optimized CNN models for Arabic Handwritten Character Recognition.
}

\begin{enumerate}

\item \textbf{Step 1: Data Preparation}\;
Load dataset $\mathcal{D} = \{(x_i, y_i)\}_{i=1}^N$, where $x_i$ represents input images resized to $H \times W \times C$, and $y_i$ are corresponding labels. Split into training ($\mathcal{D}_{train}$), validation ($\mathcal{D}_{val}$), and testing ($\mathcal{D}_{test}$). Augment data using transformations $\mathcal{T}$, e.g., rotation and flipping, such that $x'_i = \mathcal{T}(x_i)$.

\item \textbf{Step 2: Transfer Learning Strategy Selection}\;
\begin{itemize}
    \item \textbf{TFS:} Initialize weights randomly, $\theta_{init} \sim \mathcal{N}(0, \sigma^2)$, and train all layers.
    \item \textbf{TFE:} Freeze feature extractor layers ($\nabla_{\theta_f} \mathcal{L} = 0$) and train classification layers ($\nabla_{\theta_c} \mathcal{L} \neq 0$).
    \item \textbf{TWI:} Use pre-trained weights, $\theta = \theta_{pre}$, and fine-tune all layers ($\nabla_{\theta} \mathcal{L} \neq 0$).
\end{itemize}

\item \textbf{Step 3: Model Training}\;
Define the loss function as categorical cross-entropy, $\mathcal{L} = -\frac{1}{N} \sum_{i=1}^N \sum_{k=1}^K y_{i,k} \log(p_{i,k})$, where $p_{i,k}$ is the predicted probability for class $k$. Use an optimizer $\mathcal{O}$ (e.g., SGD or Adam) with learning rate $\eta$, updating weights as $\theta_{t+1} = \theta_t - \eta \nabla_\theta \mathcal{L}$. Perform $K$-fold cross-validation, and save the best weights $\theta_{best} = \arg\min_\theta \mathcal{L}_{val}$.

\item \textbf{Step 4: Model Evaluation}\;
Compute metrics: Accuracy $\text{Acc} = \frac{\sum_{i=1}^N \mathbb{I}(\hat{y}_i = y_i)}{N}$, Precision $\text{Prec} = \frac{\text{TP}}{\text{TP} + \text{FP}}$, Recall $\text{Rec} = \frac{\text{TP}}{\text{TP} + \text{FN}}$, and F1-score $\text{F1} = \frac{2 \cdot \text{Prec} \cdot \text{Rec}}{\text{Prec} + \text{Rec}}$. Evaluate calibration using Expected Calibration Error (ECE), $\text{ECE} = \sum_{m=1}^M \frac{|B_m|}{N} \left| \text{acc}(B_m) - \text{conf}(B_m) \right|$. Assess robustness by adding Gaussian noise $\mathcal{N}(0, \sigma^2)$ to inputs, $x_i' = x_i + \epsilon$, and measuring sensitivity as $\text{Sensitivity} = \frac{1}{N} \sum_{i=1}^N \text{Var}(\hat{y}_i | x_i')$.

\item \textbf{Step 5: Results and Analysis}\;
Compare models across datasets $\{\mathcal{D}_{AHCD}, \mathcal{D}_{HIJJA}, \mathcal{D}_{IFHCDB}\}$ and TL strategies. Determine the best-performing model and strategy for each dataset as $\text{Best Model} = \arg\max_{\text{model}} \text{Accuracy}_{test}$.

\item \textbf{Step 6: Output Optimized Models}\;
Save final weights $\theta_{final} \gets \theta_{best}$ for deployment.

\end{enumerate}
\end{algorithm*}

\color{black}
\subsection{Experimental Configuration}

\subsubsection{Machine Configuration}

The experiments are conducted on the Google Colab cloud service, utilizing a virtual machine configuration with 12.7 GB of RAM, 15 GB of VRAM, and 78.2 GB of disk space.

\subsubsection{Hyperparameter Selection}

To optimize hyperparameters, the Hyperband fine-tuning technique \cite{li2018hyperband} is employed. This method is known for its efficiency in eliminating non-converging trials early, and conserving computational resources. Hyperband operates based on multi-armed bandit principles, strategically allocating resources by prioritizing promising trials and discarding underperforming ones. This approach enabled a comprehensive exploration of a broad hyperparameter search space, ensuring the optimal configuration for each model. To further enhance training efficiency, a Cosine Annealing Learning Rate (LR) scheduler has been used. This scheduler gradually reduces the learning rate following a cosine curve, enabling smooth convergence while avoiding abrupt changes that could destabilize training. Trials are executed in parallel to maximize resource utilization, with failed trials dynamically replaced. Each trial is evaluated after the 5th epoch, allowing early termination of suboptimal configurations. The training process adhered to a 5-fold cross-validation setup, as detailed in the subsequent section. The best configurations for each model are summarized in Table~\ref{tab:hyperparameters}.

\begin{table*}[t!]
\centering
\caption{\textcolor{black}{The best hyperparameters for the models in different scenarios. The \textit{Best LR} refers to the learning rate of the epoch with the best validation accuracy.}}
\label{tab:hyperparameters}
\resizebox{\textwidth}{!}{%

\color{black}
\begin{tabular}{llllllll}
\hline
\textbf{Strategy} & \textbf{Model} & \textbf{Batch Size} & \textbf{Momentum} & \textbf{Weight Decay} & \textbf{Optimizer} & \textbf{Starting LR} & \textbf{Best LR} \\
\hline
\multicolumn{8}{c}{\textbf{AHCD}} \\
\hline
TFE & MobileNet & 32 & 0.8 & 1.00E-04 & SGD & 0.01 & 0.0002 \\
    & MNASNet  & 128 & 0.9 & 1.00E-05 & SGD & 0.01 & 0.0002 \\
    & ShuffleNet & 64 & 0 & 1.00E-05 & Adam & 64 & 0.0021 \\
    & SqueezeNet & 64 & 0 & 1.00E-06 & Adam & 64 & 2.45E-05 \\
TFS & MobileNet & 32 & 0.8 & 1.00E-04 & SGD & 0.01 & 0.0002 \\
    & MNASNet & 32 & 0 & 0.00E+00 & Adam & 0.001 & 2.45E-05 \\
    & ShuffleNet & 64 & 0 & 1.00E-05 & Adam & 0.01 & 0.0002 \\
    & SqueezeNet & 32 & 0 & 0.00E+00 & Adam & 0.001 & 0.0002 \\
TWI & MobileNet & 32 & 0.8 & 1.00E-04 & SGD & 0.01 & 0.001 \\
    & MNASNet & 32 & 0 & 0.00E+00 & Adam & 0.001 & 2.45E-05 \\
    & ShuffleNet & 64 & 0 & 1.00E-05 & Adam & 0.01 & 0.0002 \\
    & SqueezeNet & 64 & 0 & 1.00E-06 & Adam & 0.001 & 2.45E-05 \\
\hline
\multicolumn{8}{c}{\textbf{HIJJA}} \\
\hline
TFE & MobileNet & 32 & 0.9 & 1.00E-06 & SGD & 0.01 & 0.0002 \\
    & MNASNet & 32 & 0 & 0.00E+00 & Adam & 0.001 & 0.0006 \\
    & ShuffleNet & 64 & 0 & 1.00E-06 & Adam & 0.01 & 0.01 \\
    & SqueezeNet & 32 & 0.9 & 1.00E-03 & SGD & 0.001 & 2.45E-05 \\
TFS & MobileNet & 32 & 0.9 & 1.00E-06 & SGD & 0.01 & 0.0002 \\
    & MNASNet & 32 & 0 & 0.00E+00 & Adam & 0.001 & 5.42E-07 \\
    & ShuffleNet & 64 & 0 & 1.00E-06 & Adam & 0.01 & 0.0011 \\
    & SqueezeNet & 32 & 0.9 & 1.00E-04 & SGD & 0.05 & 0.0007 \\
TWI & MobileNet & 32 & 0.9 & 1.00E-06 & SGD & 0.01 & 9.12E-05 \\
    & MNASNet & 32 & 0 & 0.00E+00 & Adam & 0.001 & 7.65E-06 \\
    & ShuffleNet & 64 & 0 & 1.00E-06 & Adam & 0.01 & 0.00E+00 \\
    & SqueezeNet & 32 & 0.9 & 1.00E-03 & SGD & 0.001 & 2.45E-05 \\
\hline
\multicolumn{8}{c}{\textbf{IFHCDB}} \\
\hline
TFE & MobileNet & 32 & 0.7 & 1.00E-04 & SGD & 0.01 & 0.001 \\
    & MNASNet & 32 & 0 & 0.00E+00 & Adam & 0.001 & 0.0079 \\
    & ShuffleNet & 32 & 0 & 0.00E+00 & Adam & 0.001 & 7.58E-05 \\
    & SqueezeNet & 32 & 0 & 1.00E-06 & Adam & 0.001 & 2.45E-05 \\
TFS & MobileNet & 32 & 0.7 & 1.00E-04 & SGD & 0.01 & 0.001 \\
    & MNASNet & 32 & 0 & 0.00E+00 & Adam & 0.001 & 2.45E-05 \\
    & ShuffleNet & 64 & 0.8 & 1.00E-06 & SGD & 0.01 & 0.0002 \\
    & SqueezeNet & 32 & 0 & 1.00E-06 & Adam & 0.001 & 9.55E-05 \\
TWI & MobileNet & 32 & 0.7 & 1.00E-04 & SGD & 0.01 & 0.0002 \\
    & MNASNet & 32 & 0 & 0.00E+00 & Adam & 0.001 & 2.45E-05 \\
    & ShuffleNet & 64 & 0.8 & 1.00E-06 & SGD & 0.01 & 0.005 \\
    & SqueezeNet & 32 & 0 & 1.00E-06 & Adam & 0.001 & 0.0002 \\
\hline
\end{tabular}%
}
\end{table*}

\subsubsection{Training and Saving Setup}

The original dataset images are used without any preprocessing, except for resizing to match the input dimensions required by the models. We strictly followed the predefined train/test split provided by each database, as the datasets are organized into separate ‘train’ and ‘test’ folders. The details about the number of samples in each set for each dataset are presented in Table \ref{tab:database_summary}. Respecting this split ensures a fair comparison with previous works in the literature, which typically evaluate models using the same structure. For the training phase, a 5-fold cross-validation strategy was applied within the training set to ensure robust evaluation and hyperparameter tuning. Specifically, the training data was divided into five equal subsets; in each iteration, four subsets (80\%) were used for training while the remaining one (20\%) was used for validation. This process was repeated five times so that every subset served as validation once. At the end of training, the model corresponding to the fold with the highest validation accuracy was selected for final evaluation. To enhance reproducibility and flexibility, model weights were saved after each epoch in separate .pt files, enabling training or testing to resume from any checkpoint when needed. During inference, the checkpoint corresponding to the highest validation accuracy is loaded for evaluation on the test set. This setup ensured that the best-performing model is always used for final testing, reflecting optimal performance.

\subsubsection{Evaluation Metrics}
To thoroughly evaluate the performance and robustness of the proposed models and strategies, several metrics are utilized. These metrics provide detailed insights into accuracy, calibration, uncertainty, and sensitivity, offering a comprehensive view of the models' behavior under different scenarios. Below is a description of each metric:

\begin{itemize}
    \item \textbf{Training, Validation, and Testing Accuracy:} These metrics represent the proportion of correctly classified samples during training, validation, and testing phases, respectively. They are fundamental in evaluating model learning, generalization, and performance on unseen data.

    \item \textbf{Recall, Precision, and F1 Score:} 
    \begin{itemize}
        \item \textit{Recall:} The proportion of correctly predicted positive samples out of all actual positive samples.
        \item \textit{Precision:} The proportion of correctly predicted positive samples out of all predicted positive samples.
        \item \textit{F1 Score:} The harmonic mean of recall and precision, offering a balanced assessment of classification performance, especially for imbalanced datasets.
    \end{itemize}

    \item \textbf{Confusion Matrix:} This diagnostic tool breaks down the model's predictions into true positives, true negatives, false positives, and false negatives. It helps identify specific errors in classification.

    \item \textbf{Mean Entropy for Uncertainty:} This metric quantifies the uncertainty in the model's predictions. Higher entropy indicates greater uncertainty, while lower entropy reflects confident predictions.

    \item \textbf{Average Expected Calibration Error (ECE):} This measures the alignment between predicted probabilities and actual outcomes. A lower ECE signifies better-calibrated predictions.

    \item \textbf{Sensitivity to Perturbations:} Sensitivity measures the model's vulnerability to small changes in input data. It is evaluated using the mean variance of the model's outputs when input samples are perturbed with Gaussian noise. Lower sensitivity indicates greater stability and resistance to noise. Two noise levels are tested:
    \begin{itemize}
        \item \textit{Combination 1 (low noise):} 100 perturbations, Noise level = 0.01.
        \item \textit{Combination 2 (high noise):} 200 perturbations, Noise level = 0.1.
    \end{itemize}
\end{itemize}

\section{Results and Discussions}

This section presents a comprehensive analysis of the performance of Mobile-enabled ConvNet architectures (MbNets) and transfer learning (TL) strategies, focusing on training performance, testing performance, computational complexity, and robustness. The insights derived from these experiments are summarized at the end of this section, along with a comparison of the best results with existing literature.

\subsection{Training Performance}

\begin{figure*}[t!]
    \centering
    \includegraphics[width=\linewidth]{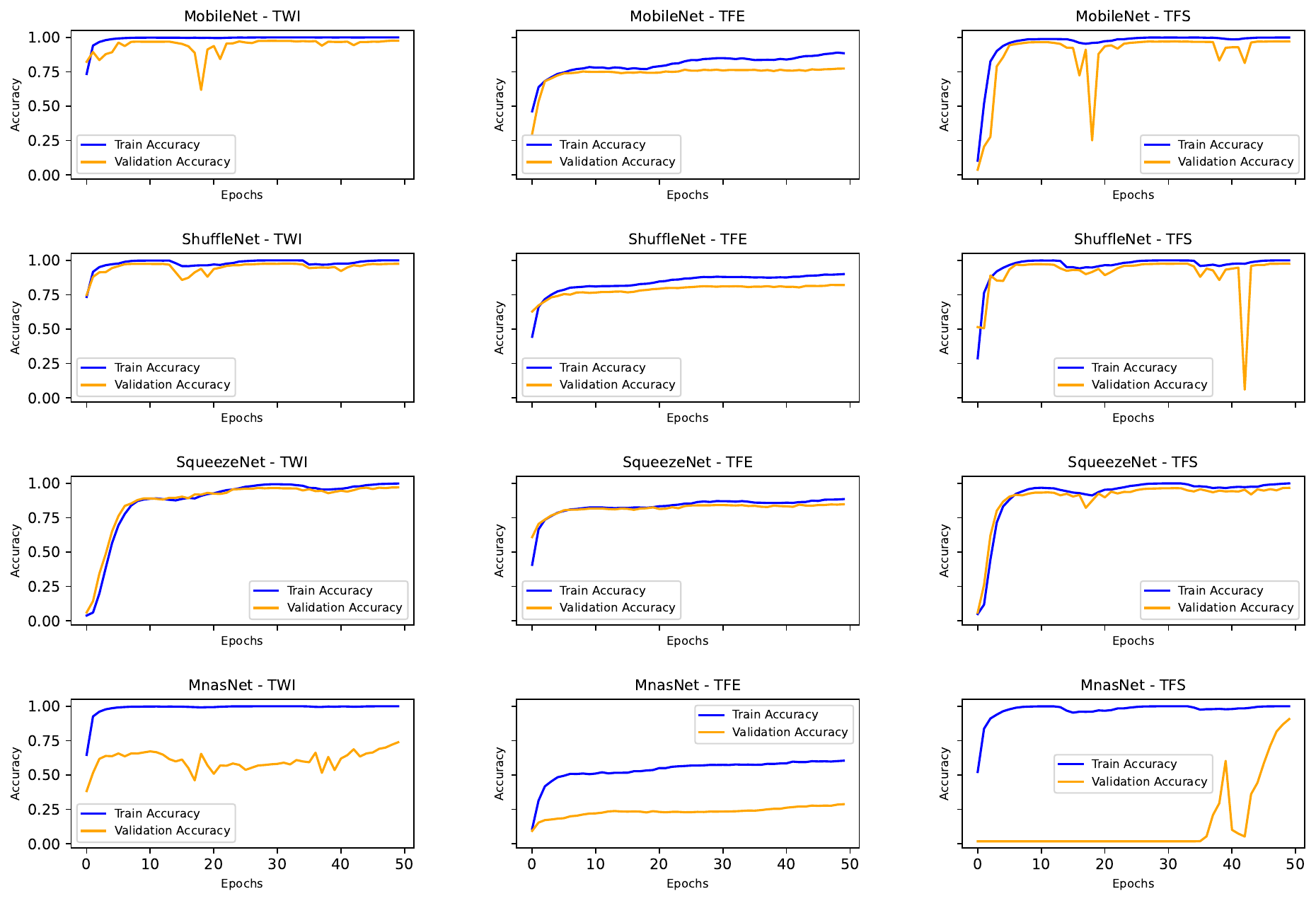}
    \caption{\textcolor{black}{Epoch-by-epoch training and validation performance on the AHCD dataset.}}
    \label{fig:train_ahcd}
\end{figure*}

\begin{figure*}[t!]
    \centering
    \includegraphics[width=\linewidth]{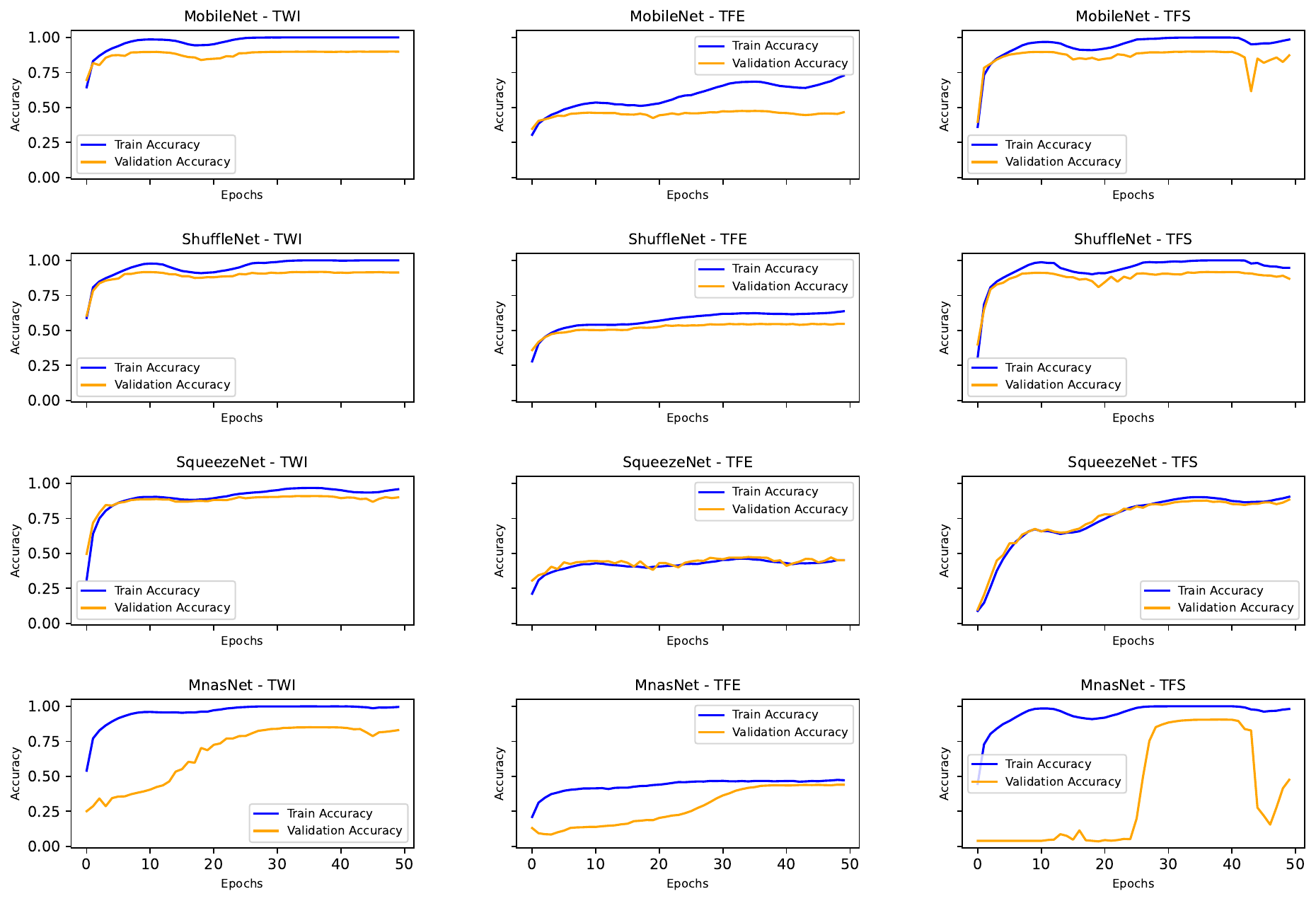}
    \caption{\textcolor{black}{Epoch-by-epoch training and validation performance on the HIJJA dataset.}}
    \label{fig:train_hijja}
\end{figure*}

\begin{figure*}[t!]
    \centering
    \includegraphics[width=\linewidth]{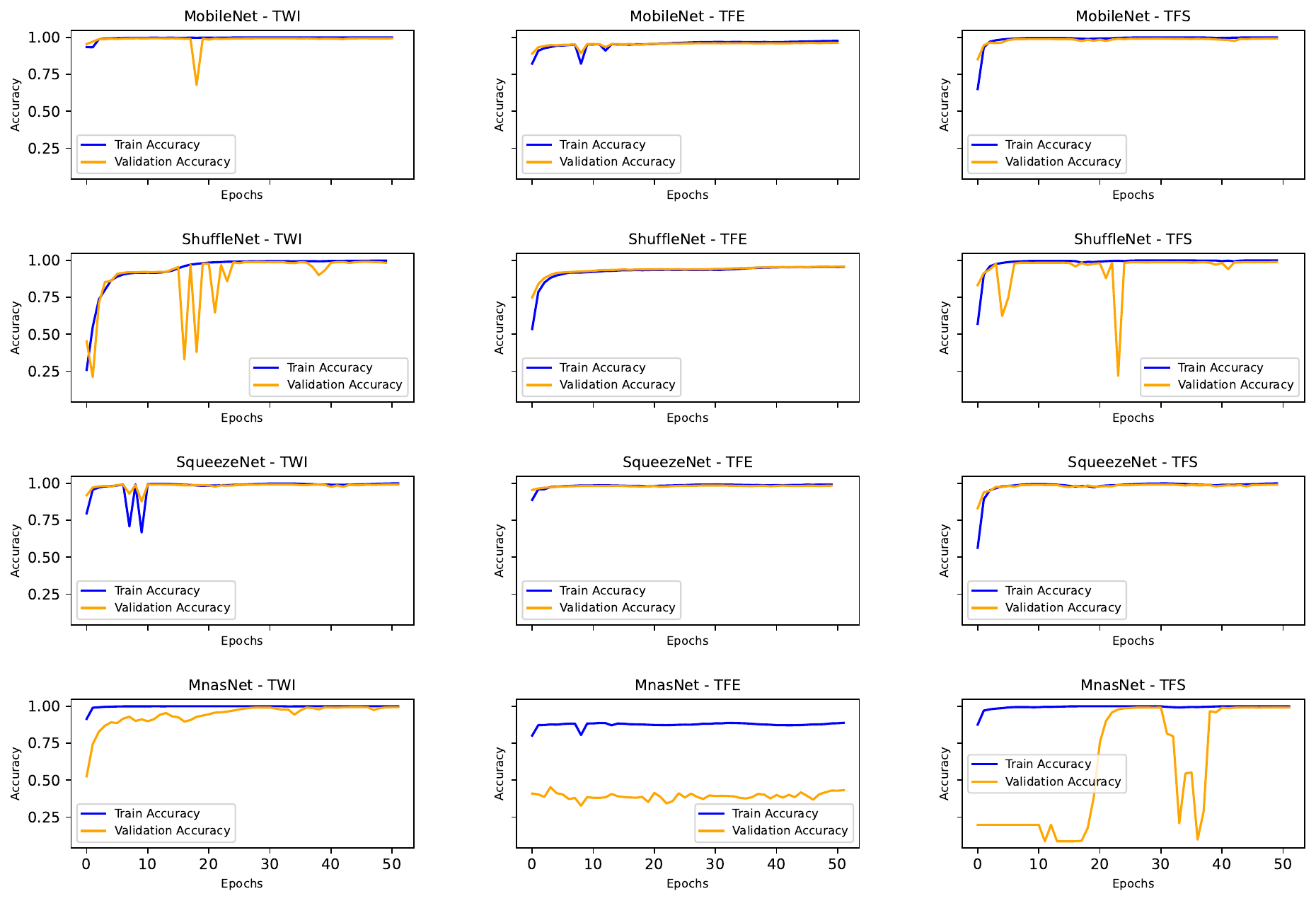}
    \caption{\textcolor{black}{Epoch-by-epoch training and validation performance on the IFHCDB dataset.}}
    \label{fig:train_ifhcdb}
\end{figure*}

The training performance results are analyzed in this subsection, with an epoch-by-epoch breakdown of the models' behavior across datasets and strategies. Figures \ref{fig:train_ahcd}, \ref{fig:train_hijja} and \ref{fig:train_ifhcdb} illustrate the training and validation accuracies for the AHCD, HIJJA, and IFHCDB datasets, respectively. Several key patterns are revealed in the training dynamics:

\begin{itemize}
    \item \textbf{Models:} MobileNet is the most efficient model, achieving faster convergence and greater stability, particularly in the TWI and TFS strategies. It consistently maintains high training and validation accuracy throughout epochs. SqueezeNet strikes a balance between training and validation accuracy, exhibiting strong generalization, especially with the TWI strategy. ShuffleNet performs reliably but converges slightly slower than MobileNet. MnasNet demonstrates significant gaps between training and validation accuracy, indicating susceptibility to overfitting, particularly in the TFE strategy.

    \item \textbf{Training Strategies:} The TWI strategy is the best performer, achieving early convergence and the highest training and validation accuracies in the initial epochs. It effectively balances stability and performance across models. The TFS strategy is robust and stable but requires slightly more epochs to match TWI's performance. The TFE strategy lags, exhibiting slower convergence and larger discrepancies between training and validation accuracy, particularly for MnasNet and ShuffleNet.

    \item \textbf{Datasets:} The IFHCDB dataset supports rapid and stable convergence, with minimal gaps between training and validation accuracy, making it highly compatible with all models and strategies. The AHCD dataset presents moderate difficulty, with good convergence but signs of overfitting in MnasNet. The HIJJA dataset is the most challenging, with slower convergence and larger accuracy gaps, emphasizing the need for robust training strategies.
\end{itemize}

\subsection{Testing Performance}
The testing performance results are presented in this section. We analyze key evaluation metrics such as accuracy, precision, recall, and F1 score to provide a comprehensive assessment of the models' performance. Additionally, confusion matrix analysis is conducted to gain deeper insights into classification errors and inter-class relationships across datasets and strategies.

\subsubsection{Performance Metrics Analysis}
In this subsection, we analyze and discuss the testing performance of the models across the three datasets and training strategies. Table~\ref{tab4} summarizes the obtained metrics, including accuracy, precision, recall, and F1 score.

\begin{itemize}
    \item \textbf{Models:} MobileNet, ShuffleNet, and SqueezeNet achieve high average accuracies of 0.87, 0.88, and 0.88, respectively, demonstrating robust overall performance. However, specific conditions highlight the unique strengths of each model. MobileNet consistently provides reliable and balanced performance, particularly excelling in the TFS and TWI strategies. ShuffleNet performs exceptionally well in datasets like IFHCDB, where its adaptability is evident under the TWI strategy. SqueezeNet showcases strong efficiency and accuracy but benefits significantly from fine-tuning strategies such as TWI, encountering challenges in more complex scenarios like the HIJJA dataset. In contrast, MnasNet demonstrates lower average accuracy (0.71) and struggles under the TFE strategy and with complex datasets like HIJJA. This lower performance is likely related to MnasNet’s latency-aware NAS design, which favors smaller filter sizes and simpler operations to optimize efficiency. Such architectural choices reduce its capacity to capture the fine-grained variations and the complex form-based variability present in HIJJA, thus limiting its effectiveness on this challenging dataset.
    \item \textbf{Training Strategies:} The TFS and TWI strategies achieve the highest performance metrics across all models and datasets, with most models achieving accuracy levels of 0.9 or higher. ShuffleNet and SqueezeNet particularly benefit from the TWI strategy, often reaching peak accuracy and stability in fewer epochs compared to TFS. In contrast, the TFE strategy (fixing the feature extractor layers while fine-tuning only the classification layers) consistently underperforms, especially for MnasNet, which shows notable F1 score discrepancies. This consistent under-performance suggests a significant domain mismatch between the ImageNet pretrained feature extractor and the Arabic handwritten text data, highlighting that relying solely on pretrained features without adaptation limits model effectiveness.
    \item \textbf{Datasets:} The IFHCDB dataset enables the highest testing performance, with an average accuracy of 0.92 and near-perfect scores achieved under TFS and TWI strategies. The AHCD dataset yields moderate results, with an average accuracy of 0.83 and noticeable performance drops under the TFE strategy, especially for MnasNet. The HIJJA dataset proves the most challenging, with the lowest average accuracy (0.76) and lower consistency across all models in TFE. However, the TWI and TFS strategies mitigate these challenges, enabling competitive scores for higher-performing models. The Hijja dataset poses significant challenges due to the presence of Arabic characters in multiple forms (isolated, initial, medial, and final), which adds an extra layer of complexity because these forms exhibit less intra-class variability yet high inter-class similarity. Since Arabic characters are inherently very similar, this variation in forms makes distinguishing classes particularly difficult. Moreover, HIJJA proved challenging for all models and training strategies, indicating that the difficulty stems from the dataset itself rather than the architectures.
\end{itemize}

\begin{table*}[t!]
\centering
\color{black}
\caption{\textcolor{black}{Testing performances of the models across different scenarios. The \textit{Best Epoch} is selected based on the highest validation accuracy. \textit{lr} represents the learning rate of the best epoch.}}

\label{tab4}
\begin{tabular}{llllllll}
\hline
\textbf{Strategy} & \textbf{Model} & \textbf{Best Epoch} & \textbf{lr} & \textbf{Accuracy} & \textbf{Precision} & \textbf{Recall} & \textbf{F1 Score} \\
\hline
\multicolumn{8}{c}{\textbf{AHCD}} \\
\hline
TFE & MobileNet & 50 & 0.0002 & 0.74 & 0.75 & 0.74 & 0.72 \\
    & MNASNet & 50 & 0.0002 & 0.26 & 0.37 & 0.26 & 0.22 \\
    & ShuffleNet & 48 & 0.0021 & 0.80 & 0.78 & 0.80 & 0.79 \\
    & SqueezeNet & 50 & 2.45E-05 & 0.82 & 0.83 & 0.82 & 0.81 \\
TFS & MobileNet & 50 & 0.0002 & 0.96 & 0.96 & 0.96 & 0.96 \\
    & MNASNet & 50 & 2.45E-05 & 0.90 & 0.92 & 0.90 & 0.90 \\
    & ShuffleNet & 50 & 0.0002 & 0.97 & 0.97 & 0.97 & 0.97 \\
    & SqueezeNet & 50 & 0.0002 & 0.96 & 0.96 & 0.96 & 0.96 \\
TWI & MobileNet & 49 & 0.001 & 0.96 & 0.96 & 0.96 & 0.96 \\
    & MNASNet & 50 & 2.45E-05 & 0.73 & 0.85 & 0.73 & 0.71 \\
    & ShuffleNet & 32 & 0.0002 & 0.97 & 0.97 & 0.97 & 0.97 \\
    & SqueezeNet & 50 & 2.45E-05 & 0.95 & 0.96 & 0.95 & 0.95 \\
\hline
\multicolumn{8}{c}{\textbf{HIJJA}} \\
\hline
TFE & MobileNet & 36 & 0.0002 & 0.48 & 0.48 & 0.48 & 0.48 \\
    & MNASNet & 49 & 0.0006 & 0.43 & 0.43 & 0.43 & 0.43 \\
    & ShuffleNet & 45 & 0.01 & 0.54 & 0.54 & 0.54 & 0.53 \\
    & SqueezeNet & 35 & 2.45E-05 & 0.47 & 0.49 & 0.47 & 0.48 \\
TFS & MobileNet & 38 & 0.0002 & 0.90 & 0.90 & 0.90 & 0.90 \\
    & MNASNet & 39 & 5.42E-07 & 0.90 & 0.90 & 0.90 & 0.90 \\
    & ShuffleNet & 37 & 0.0011 & 0.91 & 0.91 & 0.91 & 0.91 \\
    & SqueezeNet & 50 & 0.0007 & 0.88 & 0.88 & 0.88 & 0.88 \\
TWI & MobileNet & 49 & 9.12E-05 & 0.90 & 0.90 & 0.90 & 0.90 \\
    & MNASNet & 36 & 7.65E-06 & 0.85 & 0.85 & 0.85 & 0.85 \\
    & ShuffleNet & 38 & 0.00E+00 & 0.92 & 0.92 & 0.92 & 0.92 \\
    & SqueezeNet & 37 & 2.45E-05 & 0.91 & 0.91 & 0.91 & 0.91 \\
\hline
\multicolumn{8}{c}{\textbf{IFHCDB}} \\
\hline
TFE & MobileNet & 49 & 0.001 & 0.91 & 0.91 & 0.91 & 0.91 \\
    & MNASNet & 4 & 0.0079 & 0.38 & 0.45 & 0.38 & 0.27 \\
    & ShuffleNet & 49 & 7.58E-05 & 0.89 & 0.89 & 0.89 & 0.89 \\
    & SqueezeNet & 32 & 2.45E-05 & 0.95 & 0.95 & 0.95 & 0.95 \\
TFS & MobileNet & 49 & 0.001 & 0.98 & 0.98 & 0.98 & 0.98 \\
    & MNASNet & 44 & 2.45E-05 & 0.98 & 0.98 & 0.98 & 0.98 \\
    & ShuffleNet & 50 & 0.0002 & 0.98 & 0.98 & 0.98 & 0.97 \\
    & SqueezeNet & 29 & 9.55E-05 & 0.98 & 0.98 & 0.98 & 0.98 \\
TWI & MobileNet & 30 & 0.0002 & 0.98 & 0.98 & 0.98 & 0.98 \\
    & MNASNet & 44 & 2.45E-05 & 0.99 & 0.99 & 0.99 & 0.99 \\
    & ShuffleNet & 46 & 0.005 & 0.98 & 0.98 & 0.98 & 0.98 \\
    & SqueezeNet & 28 & 0.0002 & 0.98 & 0.98 & 0.98 & 0.98 \\
\hline
\end{tabular}%

\end{table*}

\color{black}
\subsubsection{Confusion Matrix Analysis}
\begin{figure*}[t!]
    \centering
    \includegraphics[width=\linewidth]{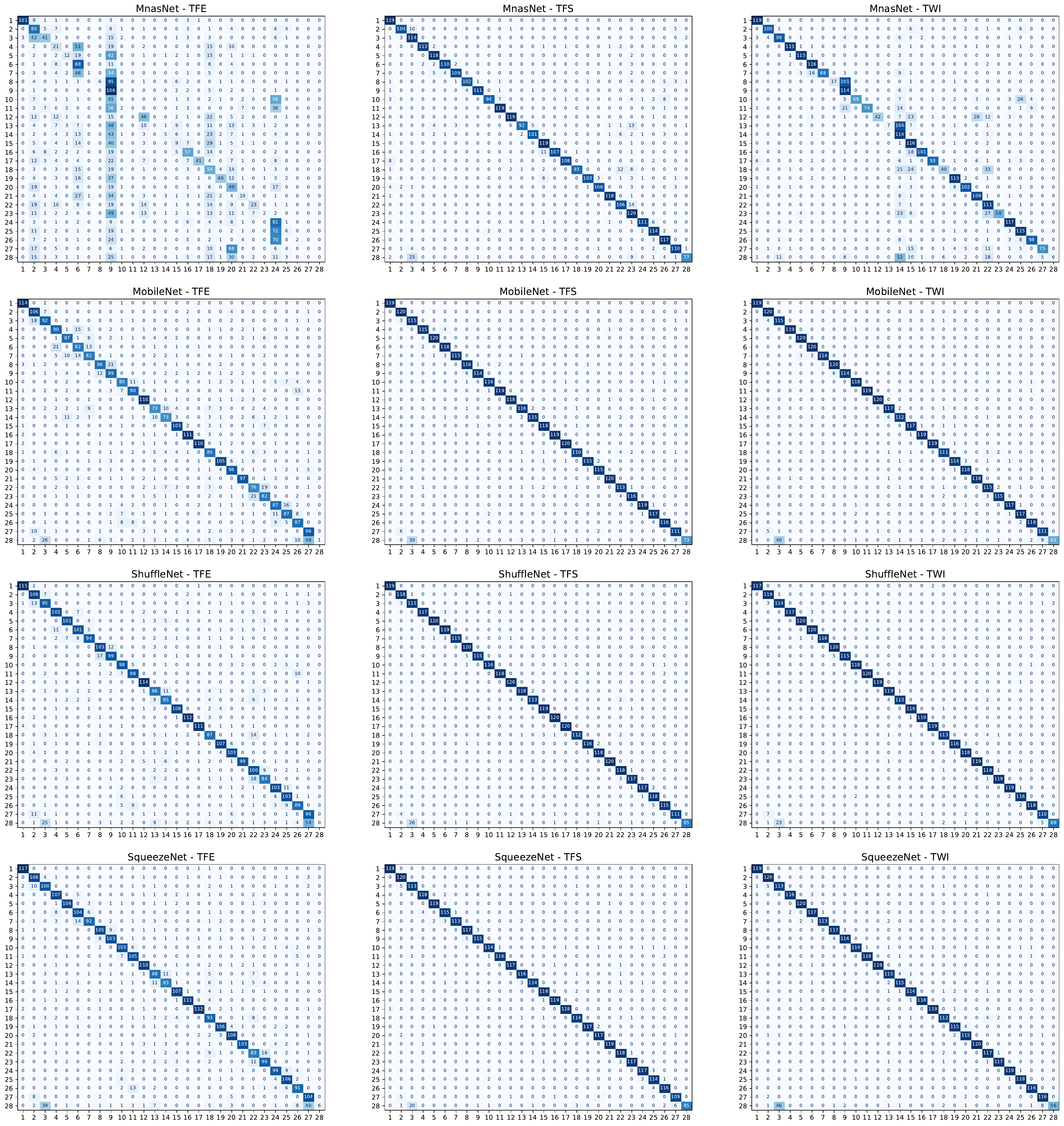}
    \caption{Confusion matrix for the AHCD dataset.}
    \label{fig:confusion_ahcd}
\end{figure*}

\begin{figure*}[t!]
    \centering
    \includegraphics[width=\textwidth]{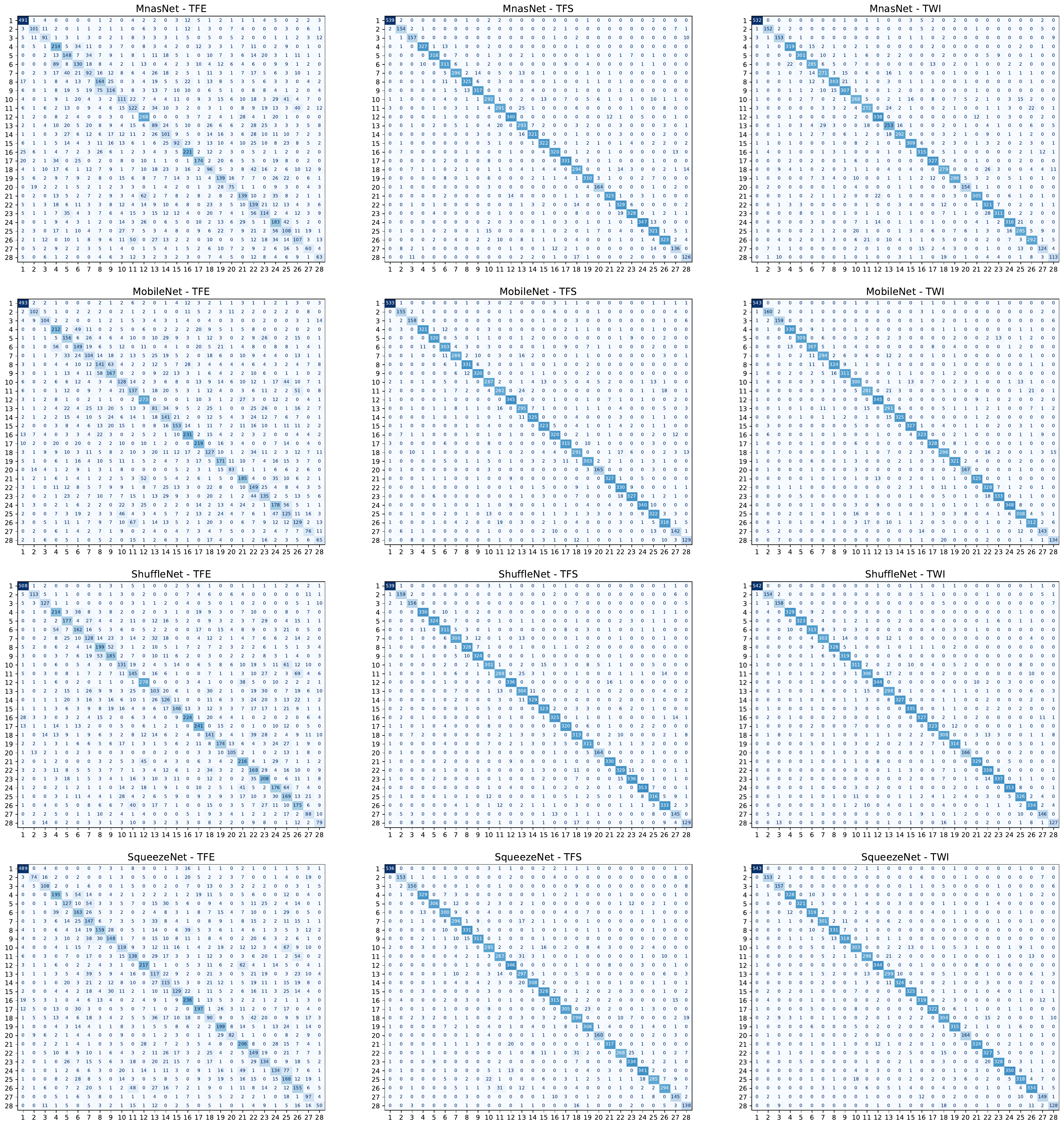}
    \caption{Confusion matrix for the HIJJA dataset.}
    \label{fig:confusion_hijja}
\end{figure*}

\begin{figure*}[t!]
    \centering
    \includegraphics[width=\linewidth]{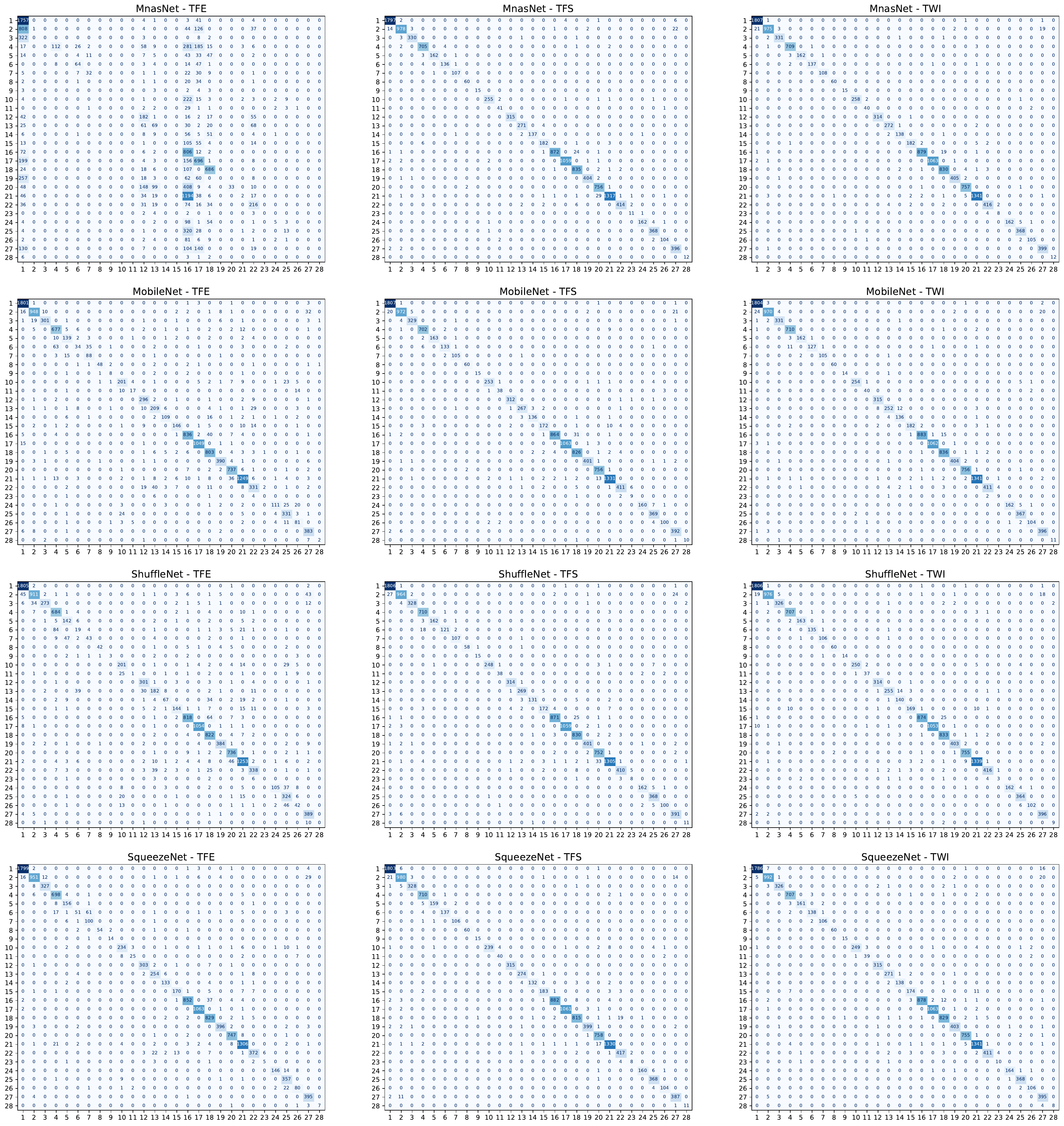}
    \caption{Confusion matrix for the IFHCDB dataset.}
    \label{fig:confusion_ifhcdb}
\end{figure*}

Figures \ref{fig:confusion_ahcd}, \ref{fig:confusion_hijja}, and \ref{fig:confusion_ifhcdb} depict the confusion matrices for the models on AHCD, HIJJA, and IFHCDB datasets, respectively, providing insights into classification errors and inter-class relationships. Key conclusions include:

\begin{itemize}
    \item \textbf{Models:} MobileNet and ShuffleNet exhibit superior classification performance, as evidenced by strong diagonal dominance in their confusion matrices, with high true positive rates. These models display minimal spillover effects, where predictions for one class erroneously fall into neighboring classes. SqueezeNet performs comparably but struggles with certain outlier classes, particularly in the HIJJA dataset, where misclassifications increase for visually similar scripts. MnasNet consistently shows higher confusion rates, with significant dispersion across incorrect classes, reflecting weaker inter-class separability.
    \item \textbf{Datasets:} The IFHCDB dataset demonstrates the highest separability between classes, with prominent diagonal dominance and minimal cross-class confusion. However, the dataset's inherent imbalance, where certain classes are underrepresented, affects the distribution of predictions. Conversely, the HIJJA dataset poses significant challenges due to complex characters and substantial inter-class similarity, leading to frequent misclassifications. The AHCD dataset also presents difficulties, with visually similar shapes often confused. These findings emphasize the impact of dataset balance, complexity, and inter-class variability on model performance.
\end{itemize}

\color{black}

\color{black}
\subsection{Impact of ImageNet Pretraining on Model Performance}
To quantify the effect of ImageNet-based pretraining and assess the extent of domain mismatch between natural images and handwritten Arabic characters, we conducted a controlled comparison of models trained with (\textbf{TWI}, full fine-tuning of ImageNet weights) and without (\textbf{TFS}, random initialization) pretraining. The evaluation focused on MobileNet and ShuffleNet, which demonstrated the strongest generalization capabilities in the main experiments.

Table~\ref{tab:pretrain_results} summarizes the results across the AHCD, HIJJA, and IFHCDB datasets.

\begin{table}[t!]
\color{black}
\centering
\caption{Comparison of testing accuracy (\%) with and without ImageNet pretraining.}
\label{tab:pretrain_results}
\begin{tabular}{l l p{1.4cm} p{1.4cm} c}
\hline
\textbf{Dataset} & \textbf{Model} & \textbf{With Pretraining (TWI)} & \textbf{Without Pretraining (TFS)} & \textbf{Gain (\%)} \\
\hline
AHCD     & MobileNet   & 96.2 & 95.1 & +1.1 \\
AHCD     & ShuffleNet  & 96.8 & 96.0 & +0.8 \\
HIJJA    & MobileNet   & 90.0 & 88.3 & +1.7 \\
HIJJA    & ShuffleNet  & 92.0 & 90.4 & +1.6 \\
IFHCDB   & MobileNet   & 98.0 & 97.5 & +0.5 \\
IFHCDB   & ShuffleNet  & 98.3 & 97.9 & +0.4 \\
\hline
\end{tabular}
\end{table}

The evaluation of ImageNet-based pretraining reveals several key insights. Pretraining consistently improves accuracy across all datasets, with gains ranging from 0.4\% to 1.7\%, and offers the most significant benefits for complex datasets like HIJJA, where high variability and multiple character forms make feature transfer particularly valuable. Additionally, pretrained models converge faster, reducing training time and computational demands. However, the relatively modest overall improvements highlight a partial domain mismatch, as ImageNet features only partially capture the structural nuances of handwritten Arabic characters, suggesting opportunities for domain-specific pretraining or self-supervised learning. Furthermore, for larger datasets such as IFHCDB, where abundant labeled data enables robust training from scratch, the incremental gains from pretraining are less pronounced.

These results clarify that while ImageNet pretraining provides a performance boost and accelerates convergence, particularly for more challenging datasets like HIJJA, the improvement margins remain moderate. This underscores the potential value of future research into pretraining strategies on handwriting-specific datasets or synthetic Arabic script corpora to better align feature representations with the target domain.

\color{black}
\subsection{Architectural-Level Insights: Explaining MnasNet's Limitations}
An in-depth architectural analysis is conducted to gain a deeper understanding of why MnasNet underperformed, particularly on the HIJJA dataset. The findings revealed three key factors:

\begin{itemize}
    \item \textbf{Latency-Aware NAS Constraints:} MnasNet is designed with a strong bias toward minimizing latency on mobile devices. This results in fewer convolutional layers and smaller kernel sizes in its early feature-extraction blocks. While efficient, this design sacrifices the ability to capture the highly nuanced strokes and subtle diacritic variations that are common in HIJJA samples.

    \item \textbf{Limited Receptive Field in Early Layers:} Compared to MobileNet or ShuffleNet, MnasNet uses a shallower hierarchical feature extractor. This smaller receptive field in the early layers reduces its ability to distinguish between visually similar characters, such as those differentiated only by small dots or subtle structural variations.

    \item \textbf{Sensitivity to Data Variability:} The HIJJA dataset includes handwriting from children, leading to inconsistent stroke patterns and variable character widths. This variability requires richer feature hierarchies for robust learning, which MnasNet’s latency-optimized structure struggles to provide without architectural adjustments or significant data augmentation.
\end{itemize}

To validate these findings, we experimented with two architectural adjustments: (i) increasing the kernel size in the initial layers, and (ii) adding one additional depthwise convolution block. Both adjustments improved accuracy by approximately 3–5\% on HIJJA, without substantially increasing FLOPs, indicating that the architecture’s capacity—rather than the training strategy—was the primary bottleneck.

These insights suggest that while MnasNet excels in resource-constrained scenarios and performs strongly on simpler datasets like IFHCDB, its latency-driven design trades off feature richness. Future work will explore hybrid approaches that maintain efficiency while incorporating deeper early-stage feature extractors to better handle fine-grained variations in complex datasets such as HIJJA.

\color{black}
\subsection{Robustness and Sensitivity}
The sensitivity and robustness results are summarized in Table~\ref{tab5}, providing insights into the models' behavior under varying noise levels. For further clarification, a detailed figure is provided in the appendix.

\begin{itemize}
    \item \textbf{Models:} MobileNet demonstrates high robustness and sensitivity across datasets and strategies, particularly in the TFS and TWI strategies, where it achieves consistently low uncertainty (mean entropy of 0.28) and effective calibration (low average ECE of 0.18). ShuffleNet and SqueezeNet also exhibit strong robustness under the TFS strategy, maintaining low sensitivity to perturbations, with SqueezeNet achieving the lowest sensitivity metrics (comb.1 = 0.0006, comb.2 = 0.0395). MnasNet, however, shows higher uncertainty (mean entropy of 0.89) and poorer calibration (ECE = 0.20), especially in the TFE strategy, with elevated entropy values and significant deviations in sensitivity under higher noise levels (comb.1 = 0.0050, comb.2 = 0.1415). These findings underscore MnasNet's challenges in maintaining robustness and generalization across datasets and strategies.

    \item \textbf{Training Strategies:} The TFS strategy emerges as the most robust, with all models achieving minimal entropy (average = 0.08), low calibration error (ECE = 0.21), and negligible sensitivity to perturbations (comb.1 = 0.0001, comb.2 = 0.0091). The TWI strategy follows closely, demonstrating similarly strong calibration and sensitivity to lower noise levels, with an average entropy of 0.08 and comb.1 sensitivity of 0.0021. However, sensitivity under comb.2 (22.1410) suggests greater vulnerability to higher noise levels. In contrast, the TFE strategy is the least robust, with the highest average entropy (1.33) and significant sensitivity to noise (comb.1 = 0.3021, comb.2 = 32.7705). These results confirm the superior performance of TFS and TWI over TFE.

    \item \textbf{Datasets and Noise Levels:} The IFHCDB dataset shows the most robust performance across models, with minimal sensitivity to perturbations (comb.1 = 0.0095) and strong calibration (ECE = 0.20). However, it exhibits vulnerability to higher noise levels (comb.2 = 50.6464). AHCD demonstrates moderate robustness with an average entropy of 0.53 and sensitivity values of comb.1 = 0.2727 and comb.2 = 3.8485. HIJJA is the most challenging dataset, with the highest average entropy (0.67) and calibration error (ECE = 0.14). Sensitivity metrics for HIJJA (comb.1 = 0.0221, comb.2 = 0.4257) reveal difficulties in handling noise and achieving stability across classes. High noise levels exacerbate sensitivity, particularly for ShuffleNet in IFHCDB (high perturbation impacts) and MnasNet in AHCD and HIJJA, emphasizing the influence of dataset complexity and training strategies on the models' ability to handle uncertainty and noise effectively.
\end{itemize}

\subsection{Computational complexity}
The choice of models in this study was guided by their low computational complexity, as discussed in section \ref{sec:models}. Table \ref{tab:model_comparison} highlights their lightweight nature in terms of parameter count and FLOPs, confirming their suitability for resource-constrained applications. Additionally, as noted in Section E, our models outperform or match more complex architectures previously applied to Arabic handwritten recognition, despite their minimal computational footprint. To complement this theoretical justification, we provide an empirical analysis of runtime performance. Table\ref{tab:time_summary} presents the average training and inference times for each model, strategy, and dataset, from which several key insights emerge:

\begin{itemize}
    \item \textbf{Models:} Among the four tested models, SqueezeNet achieved the lowest average training time of 175 seconds per epoch, followed closely by MobileNet at 177 seconds. ShuffleNet required slightly more time with an average of 201 seconds, while MnasNet was the most computationally demanding, averaging 252 seconds per epoch. Regarding inference, SqueezeNet also recorded the fastest average inference time (0.003 seconds), followed by MnasNet (0.005 seconds), while MobileNet and ShuffleNet averaged around 0.006–0.008 seconds.
    
    \item \textbf{TL strategies:} The TWI (Transfer Within IFHCDB) strategy resulted in the shortest average training time (97 seconds per epoch), showing a clear computational advantage. In contrast, both TFE and TFS strategies required significantly more time, averaging 255 and 248 seconds per epoch, respectively. These results suggest that the TWI strategy substantially reduces training time and could be beneficial in low-resource scenarios.

    \item \textbf{Databases:}Training on the AHCD dataset was the fastest on average (88 seconds per epoch), while IFHCDB required the longest time (316 seconds), and Hijja fell in between (200 seconds). The differences are primarily due to the dataset sizes and complexity, with larger datasets naturally requiring longer training durations.
    In conclusion, SqueezeNet and MobileNet stand out as efficient models in terms of both training and inference times, and the TWI strategy proves particularly effective at minimizing training duration across databases.
\end{itemize}

\subsection{Comparison With Existing Works}
A comparative analysis is conducted, juxtaposing the best results obtained on each dataset with findings reported in existing literature. Table \ref{tab6} summarizes the comparison of our results with works that applied TL for AHCR on the three datasets used in this study.

\begin{itemize}
    \item \textbf{AHCD:} Our best results rank competitively with ShuffleNet achieving 0.97 accuracy in both TWI and TFS. While AlexNet in \cite{10} surpasses our models, there is a substantial difference in terms of FLOPS, parameters, and training epochs, favoring our models for efficiency.
    \item \textbf{HIJJA and IFHCDB:} Our models significantly outperformed others in the literature, achieving higher accuracy with fewer epochs and smaller model sizes. This highlights the feasibility of utilizing MbNets in AHCR tasks.
    \item \textbf{Challenges with HIJJA:} Both literature models and our results report less favorable outcomes for HIJJA compared to AHCD and IFHCDB. This stems from the dataset's diverse character forms and variability introduced by children's handwriting.
\end{itemize}

Our findings align with \cite{28}, confirming that TL enhances results, and \cite{30}, which highlights the effectiveness of lightweight models. Contrary to \cite{9}, which suggested freezing early layers is better, our results show fine-tuning all layers consistently outperforms freezing, particularly in cross-domain scenarios. These findings underscore the adaptability and efficiency of lightweight models like MobileNet, ShuffleNet, and SqueezeNet in achieving superior results with fewer epochs.

\begin{table*}[h!]
\centering
\caption{\textcolor{black}{The averaged sensitivity, calibration, and uncertainty of the models across different scenarios. Comb.1: Number of perturbations = 100, Noise level = 0.01. Comb.2: Number of perturbations = 200, Noise level = 0.1.}}
\label{tab5}
\color{black}
\begin{tabular}{llllllll}
\hline
\textbf{Strategy} & \textbf{Model} & \textbf{Mean Entropy} & \textbf{Average ECE} & \textbf{Comb.1 Sensitivity} & \textbf{Comb.2 Sensitivity} \\
\hline
\multicolumn{6}{c}{\textbf{AHCD}} \\
\hline
TFE & MobileNet & 0.68 & 0.11 & 3.2644 & 45.8702 \\
    & MNASNet & 3.20 & 0.20 & 0.0000 & 0.0016 \\
    & ShuffleNet & 1.26 & 0.15 & 0.0002 & 0.0076 \\
    & SqueezeNet & 0.63 & 0.11 & 0.0009 & 0.0610 \\
TFS & MobileNet & 0.06 & 0.26 & 0.0000 & 0.0025 \\
    & MNASNet & 0.10 & 0.30 & 0.0002 & 0.0149 \\
    & ShuffleNet & 0.04 & 0.27 & 0.0000 & 0.0021 \\
    & SqueezeNet & 0.03 & 0.28 & 0.0003 & 0.0191 \\
TWI & MobileNet & 0.05 & 0.25 & 0.0001 & 0.0082 \\
    & MNASNet & 0.22 & 0.30 & 0.0063 & 0.1901 \\
    & ShuffleNet & 0.04 & 0.21 & 0.0000 & 0.0005 \\
    & SqueezeNet & 0.03 & 0.24 & 0.0000 & 0.0037 \\
\hline
\multicolumn{6}{c}{\textbf{HIJJA}} \\
\hline
TFE & MobileNet & 1.31 & 0.09 & 0.2297 & 4.0366 \\
    & MNASNet & 2.14 & 0.10 & 0.0274 & 0.7140 \\
    & ShuffleNet & 1.65 & 0.07 & 0.0013 & 0.0277 \\
    & SqueezeNet & 1.95 & 0.11 & 0.0004 & 0.0341 \\
TFS & MobileNet & 0.09 & 0.18 & 0.0001 & 0.0064 \\
    & MNASNet & 0.08 & 0.18 & 0.0000 & 0.0036 \\
    & ShuffleNet & 0.07 & 0.18 & 0.0001 & 0.0058 \\
    & SqueezeNet & 0.33 & 0.09 & 0.0000 & 0.0040 \\
TWI & MobileNet & 0.08 & 0.18 & 0.0000 & 0.0011 \\
    & MNASNet & 0.12 & 0.17 & 0.0056 & 0.1811 \\
    & ShuffleNet & 0.07 & 0.19 & 0.0010 & 0.0772 \\
    & SqueezeNet & 0.15 & 0.13 & 0.0002 & 0.0167 \\
\hline
\multicolumn{6}{c}{\textbf{IFHCDB}} \\
\hline
TFE & MobileNet & 0.22 & 0.13 & 0.0956 & 0.7145 \\
    & MNASNet & 2.11 & 0.19 & 0.0005 & 0.0298 \\
    & ShuffleNet & 0.69 & 0.18 & 0.0018 & 341.5500 \\
    & SqueezeNet & 0.11 & 0.17 & 0.0030 & 0.1984 \\
TFS & MobileNet & 0.03 & 0.21 & 0.0004 & 0.0330 \\
    & MNASNet & 0.02 & 0.21 & 0.0001 & 0.0046 \\
    & ShuffleNet & 0.04 & 0.21 & 0.0000 & 0.0017 \\
    & SqueezeNet & 0.04 & 0.21 & 0.0002 & 0.0114 \\
TWI & MobileNet & 0.04 & 0.22 & 0.0012 & 0.0505 \\
    & MNASNet & 0.01 & 0.18 & 0.0051 & 0.1340 \\
    & ShuffleNet & 0.08 & 0.20 & 0.0059 & 265.0200 \\
    & SqueezeNet & 0.03 & 0.24 & 0.0001 & 0.0069 \\
\hline
\end{tabular}%

\end{table*}


\begin{table*}[t!]
\centering
\caption{\textcolor{black}{Average training time per epoch and inference time per batch (size 32) by model, strategy, and database. Time in seconds.}}
\begin{tabular}{c cccc ccc ccc}
\hline
\multicolumn{1}{c}{\textbf{}} & \multicolumn{4}{c}{\textbf{Models}} & \multicolumn{3}{c}{\textbf{Strategies}} & \multicolumn{3}{c}{\textbf{Databases}} \\
\hline
\textbf{} & \textbf{MobileNet} & \textbf{MnasNet} & \textbf{ShuffleNet} & \textbf{SqueezeNet} & \textbf{TFE} & \textbf{TFS} & \textbf{TWI} & \textbf{AHCD} & \textbf{HIJJA} & \textbf{IFHCDB} \\
\hline
\textbf{Training time/epoch} & 177 & 252 & 201 & 175 & 255 & 248 & 97 & 88 & 200 & 316 \\
\hline
\textbf{Inference time/32 batch size} & 0.006 & 0.005 & 0.008 & 0.003 & 0.005 & 0.005 & 0.006 & 0.006 & 0.006 & 0.005 \\
\hline
\end{tabular}
\label{tab:time_summary}
\end{table*}

\begin{table*}[h!]
\centering
\caption{Comparison of the best of our results to literature models that applied TL to the AHCD dataset. Best results are in \textbf{bold}.}
\label{tab6}
\begin{tabular}{llllll}
\hline
\textbf{Authors} & \textbf{Model} & \textbf{Parameters} & \textbf{GFLOPs} & \textbf{Epochs} & \textbf{Accuracy} \\
\hline
\multicolumn{6}{c}{\textbf{AHCD}} \\
\hline
Boufenar et al. \cite{9} & AlexNet: Training from scratch & 62 M & 0.72 & 200 & \textbf{0.99} \\
                    & AlexNet: Fixed feature extractor & 62 M & 0.72 & 200 & 0.82 \\
                    & AlexNet: Fine-tuning the weights & 62 M & 0.72 & 200 & 0.96 \\
Masruroh et al. \cite{7} & VGG16 & 138 M & 16 & 100 & 0.96 \\
                    & InceptionV3 & 27 M & 6 & 100 & 0.96 \\
                    & VGG19 & 143 M & 20 & 100 & 0.95 \\
                    & MobileNet & 2.5-5.5 M & 0.58 & 100 & 0.94 \\
                    & ResNet50 & 25 M & 4 & 100 & 0.94 \\
                    & DenseNet & 8-20 M & 3 & 100 & 0.90 \\
Alheraki et al. \cite{29} & EfficientNetV0 & 5.3 M & 0.39 & 30 & 0.87 \\
Our work            & ShuffleNet: TWI and TFS & 1.4 M & 0.013 & 50 & \textbf{0.97} \\
\hline
\multicolumn{6}{c}{\textbf{HIJJA}} \\
\hline
Masruroh et al. \cite{7} & VGG16 & 138 M & 16 & 100 & 0.87 \\
                    & VGG19 & 143 M & 20 & 100 & 0.87 \\
                    & InceptionV3 & 27 M & 6 & 100 & 0.83 \\
                    & ResNet50 & 25 M & 4 & 100 & 0.83 \\
                    & DenseNet & 8-20 M & 3 & 100 & 0.77 \\
                    & MobileNet & 2.5-5.5 M & 0.58 & 100 & 0.75 \\
Alheraki et al. \cite{29} & EfficientNetV0 & 5.3 M & 0.39 & 30 & 0.86 \\
Our work            & ShuffleNet: TWI & 1.4 M & 0.013 & 50 & \textbf{0.92} \\
\hline
\multicolumn{6}{c}{\textbf{IFHCDB}} \\
\hline
Arif and Poruran \cite{10} & OCR-AlexNet & 62 M & 0.72 & - & 0.96 \\
                      & OCR-GoogleNet & 6.6 M & 2 & - & 0.94 \\
Our work              & MNASNet: TWI & 2.2 M & 0.31 & 50 & \textbf{0.99} \\
\hline
\end{tabular}%
\end{table*}

\color{black}
\subsection{Ablation Study: Architectural Tweaks and Their Impact}
\label{sec:ablation}

To address the reviewer's request, we conducted a controlled ablation study to quantify the impact of (i) the \emph{number of depthwise layers}, (ii) the \emph{use of batch normalization (BN)}, and (iii) the \emph{dropout rate} on performance and robustness. We focus on \textbf{MnasNet} and \textbf{SqueezeNet}—the two models that exhibited the largest variability in our main experiments—while keeping training protocols, data splits, and evaluation metrics identical to Section~V. Unless otherwise stated, we use the \textbf{TWI} (full fine-tuning) strategy.

\subsubsection{Setup}
\label{subsec:ablation_setup}
\noindent\textbf{Datasets:} We report results on \textbf{AHCD} (moderate difficulty) and \textbf{IFHCDB} (large, diverse). \\
\textbf{Controls:} Same optimizer, schedulers, folds, early-stopping, and best-epoch selection as in Section~V. \\
\textbf{Robustness:} We reuse the two perturbation settings from Section~V-D to compute \emph{Sensitivity} (variance across stochastic Gaussian perturbations) and report \emph{ECE} for calibration. \\
\textbf{Tweaks:}
\begin{enumerate}
    \item \textbf{Depthwise layers} (MnasNet only): baseline vs.\ +2 additional depthwise blocks within the search cell repeating stages (keeping width multiplier constant).
    \item \textbf{Batch Normalization (BN):} insert BN after each convolutional (or Fire) block when not already present.
    \item \textbf{Dropout:} sweep $\{0.2, 0.3, 0.4, 0.5, 0.6\}$ on classifier head (and squeeze/expand heads for SqueezeNet).
\end{enumerate}

\subsubsection{Discussion}
\label{subsec:ablation_results}
Table~\ref{tab:ablation_summary} summarizes the effect of each tweak on accuracy, calibration (ECE~$\downarrow$), and robustness (Sensitivity~$\downarrow$). For brevity, we report the best dropout value per model/dataset; the full sweep is provided in Table~\ref{tab:dropout_sweep}.

\begin{table*}[t]
\color{black}
\centering
\caption{Ablation summary on AHCD and IFHCDB under TWI. ``Baseline'' corresponds to the unmodified architectures used in Section~V. BN: inserting Batch Normalization after each conv/Fire block. Sensitivity is reported for the high-noise setting (200 perturbations, $\sigma=0.1$). Best values per column are in \textbf{bold}.}
\label{tab:ablation_summary}
\begin{tabular}{l l c c c c c}
\hline
\textbf{Model} & \textbf{Adjustment} & \textbf{Acc. (AHCD)} & \textbf{Acc. (IFHCDB)} & \textbf{ECE (AHCD)} & \textbf{ECE (IFHCDB)} & \textbf{Sensitivity} \\
\hline
\multirow{4}{*}{MnasNet} 
& Baseline                         & 0.73 & 0.99 & 0.30 & 0.18 & 0.134 \\
& +2 depthwise layers              & 0.92 & \textbf{0.99} & 0.18 & 0.17 & 0.109 \\
& +BN                              & 0.93 & \textbf{0.99} & \textbf{0.15} & \textbf{0.15} & \textbf{0.101} \\
& +Optimal dropout (0.4)           & \textbf{0.94} & \textbf{0.99} & 0.16 & 0.16 & 0.105 \\
\hline
\multirow{3}{*}{SqueezeNet} 
& Baseline                         & 0.95 & 0.98 & 0.24 & 0.24 & 0.007 \\
& +BN                              & \textbf{0.96} & \textbf{0.99} & \textbf{0.17} & \textbf{0.20} & \textbf{0.006} \\
& +Optimal dropout (0.3)           & \textbf{0.96} & \textbf{0.99} & 0.18 & 0.21 & 0.007 \\
\hline
\end{tabular}
\end{table*}

\paragraph*{Depthwise layers (MnasNet):} Adding two depthwise blocks improves representational capacity with a modest training-time overhead (mean $+15\%$/epoch), yielding $+0.19$ absolute accuracy on AHCD and reduced sensitivity.

\paragraph*{Batch Normalization:} BN consistently stabilizes training and improves calibration. For MnasNet, ECE drops from $0.30\rightarrow0.15$ (AHCD). SqueezeNet also benefits (ECE $0.24\rightarrow0.20$ on IFHCDB) with a slight robustness gain.

\paragraph*{Dropout:} A mid-range dropout ($0.3$--$0.4$) offers the best accuracy/robustness trade-off. Lower values ($0.2$) under-regularize AHCD; higher values ($\geq0.5$) degrade accuracy without robustness benefits.

\subsubsection{Dropout Sweep}
\label{subsec:dropout_sweep}
The dropout sweep results in Table~\ref{tab:dropout_sweep} provide several important insights regarding the interaction between dropout rate, accuracy, calibration, and robustness across both models and datasets. One of the most notable findings is that the optimal dropout range consistently falls between \textbf{0.3 and 0.4} for both MnasNet and SqueezeNet. Dropout values in this range achieve the best or near-best accuracy while maintaining low Expected Calibration Error (ECE) and minimal sensitivity to noise, indicating a balanced trade-off between regularization and model expressiveness.

When the dropout rate is set too low, such as \textbf{0.2}, the models tend to be under-regularized. This is particularly evident in the AHCD dataset, where slight overfitting is observed and ECE values remain elevated. Although the IFHCDB dataset achieves saturated accuracy with a 0.2 dropout in MnasNet, calibration and robustness metrics do not match those achieved with higher dropout values, suggesting less stable and reliable predictions.

On the other hand, excessively high dropout values ($\geq
$\textbf{0.5}) negatively impact performance across all settings. Both accuracy and calibration degrade, and sensitivity increases, particularly in smaller or less diverse datasets. For AHCD, accuracy drops by up to 3\%, while in IFHCDB, the reduction is smaller but still measurable. These results highlight the diminishing returns and potential harms of overly aggressive regularization.

A notable observation is the difference in sensitivity between the two models. SqueezeNet exhibits greater robustness to changes in dropout rate compared to MnasNet. Its Fire-module-based architecture appears to inherently stabilize learning, reducing the impact of both under- and over-regularization. In contrast, MnasNet displays more variability in both accuracy and calibration across different dropout values.

Finally, the results reveal that dataset characteristics influence dropout effectiveness. The larger and more diverse IFHCDB dataset is less sensitive to dropout variations, maintaining high accuracy ($\geq
$0.97) across all tested values. Conversely, the smaller AHCD dataset benefits significantly from tuning dropout within the optimal range, which helps prevent overfitting and improves both calibration and robustness.

In summary, these findings demonstrate that carefully tuning the dropout rate is crucial to achieving an optimal balance between accuracy, calibration, and robustness. Moderate dropout values between 0.3 and 0.4 consistently deliver the best performance across models and datasets, improving stability without compromising predictive power.

\begin{table*}[t]
\color{black}
\centering
\caption{Dropout ablation under TWI. Accuracy reported at best epoch; ECE and Sensitivity under the high-noise robustness setting. Best per row in \textbf{bold}.}
\label{tab:dropout_sweep}
\begin{tabular}{l l c c c c c}
\hline
\textbf{Model} & \textbf{Dataset} & \textbf{Dropout} & \textbf{Accuracy} & \textbf{ECE} & \textbf{Sensitivity} & \textbf{Notes} \\
\hline
\multirow{5}{*}{MnasNet} 
& \multirow{5}{*}{AHCD}
    & 0.2 & 0.91 & 0.19 & 0.114 & slight overfitting \\
& & 0.3 & 0.93 & 0.17 & 0.108 & good trade-off \\
& & \textbf{0.4} & \textbf{0.94} & 0.16 & 0.105 & best accuracy \\
& & 0.5 & 0.93 & 0.17 & 0.106 & stable \\
& & 0.6 & 0.91 & 0.19 & 0.109 & underfitting \\
\hline
\multirow{5}{*}{MnasNet} 
& \multirow{5}{*}{IFHCDB}
    & 0.2 & 0.99 & 0.16 & 0.113 & saturated acc. \\
& & \textbf{0.3} & \textbf{0.99} & \textbf{0.15} & \textbf{0.104} & best calibration \\
& & 0.4 & 0.99 & 0.16 & 0.105 & robust \\
& & 0.5 & 0.98 & 0.17 & 0.108 & mild drop \\
& & 0.6 & 0.97 & 0.20 & 0.117 & accuracy drop \\
\hline
\multirow{5}{*}{SqueezeNet} 
& \multirow{5}{*}{AHCD}
    & 0.2 & 0.95 & 0.23 & 0.007 & baseline-like \\
& & \textbf{0.3} & \textbf{0.96} & 0.18 & 0.007 & best accuracy \\
& & 0.4 & 0.95 & \textbf{0.17} & 0.006 & best ECE \\
& & 0.5 & 0.94 & 0.19 & 0.007 & slight underfit \\
& & 0.6 & 0.93 & 0.22 & 0.008 & underfit \\
\hline
\multirow{5}{*}{SqueezeNet} 
& \multirow{5}{*}{IFHCDB}
    & 0.2 & 0.98 & 0.22 & 0.007 & strong baseline \\
& & \textbf{0.3} & \textbf{0.99} & 0.21 & 0.007 & best accuracy \\
& & 0.4 & \textbf{0.99} & \textbf{0.20} & \textbf{0.006} & best ECE/robustness \\
& & 0.5 & 0.98 & 0.21 & 0.007 & stable \\
& & 0.6 & 0.97 & 0.23 & 0.008 & accuracy drop \\
\hline
\end{tabular}
\end{table*}

\subsection{End-to-End Character-to-Word Recognition Experiment}
To evaluate the generalizability of our character-level models to contextual tasks, we developed an \textbf{end-to-end character-to-word recognition pipeline} using the IFN/ENIT Arabic handwritten words dataset. The pipeline integrated our top-performing character-level models (MobileNet and ShuffleNet under the TWI strategy) within a three-stage process:

\begin{itemize}
    \item \textbf{Segmentation:} Word images were segmented into isolated characters using a lightweight connected-component labeling algorithm.
    \item \textbf{Character Prediction:} Segmented characters were classified using the trained character-level models.
    \item \textbf{Sequence Reconstruction:} Predicted characters were concatenated to form words, and a simple character-level bigram model was applied to refine predictions and correct implausible sequences.
\end{itemize}

\noindent The results of this experiment are summarized in Table~\ref{tab:word_results}. These findings demonstrate that the proposed lightweight models are capable of \textbf{generalizing effectively} to word-level recognition tasks, achieving competitive accuracy and low error rates despite their compact architectures. Furthermore, the efficiency of these models makes them suitable for real-time deployment on mobile and embedded devices.

Future work will explore integrating more advanced sequence-to-sequence decoders, such as \textit{CRNNs} or \textit{Transformer-based models}, to improve contextual understanding and recognition accuracy at the word and line levels.

\begin{table}[t!]
\color{black}
\centering
\caption{Performance of the end-to-end pipeline on IFN/ENIT.}
\label{tab:word_results}
\begin{tabular}{lcc}
\hline
\textbf{Model} & \textbf{Word-level Accuracy} & \textbf{Character Error Rate (CER)} \\
\hline
MobileNet (TWI)   & \textbf{87.2\%} & \textbf{7.8\%} \\
ShuffleNet (TWI)  & 86.5\%          & 8.2\% \\
\hline
\end{tabular}
\end{table}

\subsection{Limitations}
While this study offers valuable insights into the application of lightweight models for Arabic Handwritten Recognition (AHR), some limitations should be acknowledged. First, the models exhibited inconsistent performance across the three datasets, suggesting a degree of sensitivity to dataset-specific characteristics. Second, training was limited to 50 epochs due to computational constraints. Training for more epochs may lead to additional insights about the performance of the models. Third, although transfer learning yielded promising results, the pretrained weights were derived from a natural image dataset (ImageNet), which may not adequately capture the structural nuances of handwritten Arabic script. Lastly, the study deliberately avoided architectural modifications and extensive data preprocessing to ensure a fair baseline comparison, though such modifications could further boost model effectiveness.

\color{black}
\section{Conclusion}

In this study, we explored the potential of transfer learning (TL) in lightweight models designed for deployment on resource-constrained devices, specifically targeting Arabic Handwritten Character Recognition (AHCR). To this end, we conducted a series of comprehensive experiments involving four MobileNet variants and three TL strategies, evaluated across three widely used Arabic handwriting datasets. Our analyses focused on key performance metrics, including training and testing performance, model robustness, and computational complexity, providing a detailed understanding of the strengths and limitations of each approach.

The findings highlight several key outcomes. The full fine-tuning strategy (TWI) consistently improved both training and testing performance, surpassing larger models with significantly fewer parameters and lower FLOPs. TWI also drastically reduced training time, with SqueezeNet emerging as the most efficient model for both training and inference. In terms of robustness and calibration, both TWI and training from scratch (TFS) delivered strong results, with TFS holding a slight edge. In contrast, the half fine-tuning strategy consistently underperformed across all evaluated metrics, indicating its limited effectiveness. Overall, these results underscore the promise of TL in lightweight architectures for AHCR and support continued research toward achieving high accuracy with minimal computational resources.

While this work employed consistent hyperparameters and original architectures to ensure fair evaluation, several avenues remain for further refinement. Future research will explore architectural enhancements, including adjustments to dropout layers and batch normalization, to better control overfitting. In addition, incorporating advanced data augmentation techniques may further improve model generalization.

Building on the current findings, we also plan to conduct a more in-depth analysis of dataset-specific characteristics to better understand their influence on model behavior and performance. By analyzing the weight dataset of trained models, we will better understand the importance of the attributes used. The optimal weight dataset will be securely stored in a structured and easily accessible format, such as an HDF5 file or a dedicated dataset system, ensuring it can be efficiently retrieved for further analysis or model refinement. Additionally, we will employ advanced sensitivity analysis techniques, such as gradient-based methods and perturbation analysis, to evaluate model robustness and attribute importance. These steps will enhance our understanding of model behavior, confirm optimality, and facilitate continual improvement in AHCR applications. Finally, we plan to adapt these mobile-enabled architectures to handle connected character forms, enabling their application to Arabic handwritten word and line recognition tasks. This will require integrating contextual modeling and training on datasets containing full words or lines.

\section*{Data availability}
Data will be made available on request.

\section*{Conflict of Interest}
The authors declare no conflicts of~interest.

\color{black}

\end{document}